%% file: main.tex
\documentclass[letterpaper, 10 pt, journal, twoside]{IEEEtran}
\input{preamble.tex}

\title{\bf Learning Reactive Dexterous Grasping via Hierarchical Task-Space RL Planning and Joint-Space QP Control}

\author{Ho Jae Lee$^{1}$, Yonghyeon Lee$^{1}$, Alexander Alexiev$^{1}$, Tzu-Yuan Lin$^{1}$, Se Hwan Jeon$^{1}$, and Sangbae Kim$^{1}$

\thanks{$^{1}$Department of Mechanical Engineering, Massachusetts Institute of Technology, Cambridge, MA 02139, USA
{\tt\footnotesize(e-mail: hjlee201@mit.edu; yhl@mit.edu; aalexiev@mit.edu; tzuyuan@mit.edu; sehwan@mit.edu; sangbae@mit.edu).}}%
}

\AtBeginBibliography{\footnotesize}

\begin{document}

\maketitle
\input{commands}

\input{Sections/00_abstract}
\input{Sections/01_intro_v2}
\input{Sections/02_related_works_v2}
\input{Sections/03_preliminaries_v2}

\input{Sections/04_system_architecture}
\input{Sections/05_rl_planning}
\input{Sections/06_qp_control}
\input{Sections/07_results}
\input{Sections/08_analysis}
\input{Sections/09_conclusion}

\addtolength{\textheight}{0cm}
\footnotesize{\printbibliography}
\clearpage
\input{Sections/10_appendix}

\end{document}

%% file: preamble.tex
\IEEEoverridecommandlockouts             

\usepackage{graphics} 
\usepackage{epsfig} 
\usepackage{times} 
\usepackage{amsmath} 
\usepackage{amssymb}  
\usepackage[per-mode=symbol]{siunitx}
\usepackage{hyperref}
\usepackage{algorithm}
\usepackage{algpseudocode}
\usepackage{listings}
\usepackage[font=footnotesize]{caption}
\usepackage{subcaption}
\usepackage[isbn=false,
    backend=biber,
    style=ieee,
    bibstyle=ieee,  
    sortcites=true,   
    giveninits=true,            
    backref=false]{biblatex}
\usepackage{xcolor}
\usepackage{color,soul}
\usepackage{booktabs}
\usepackage{multirow}
\usepackage{tabularx}
\definecolor{codegreen}{rgb}{0,0.6,0}
\definecolor{codegray}{rgb}{0.5,0.5,0.5}
\definecolor{codepurple}{rgb}{0.58,0,0.82}
\definecolor{backcolour}{rgb}{0.95,0.95,0.92}

\lstdefinestyle{mystyle}{
  backgroundcolor=\color{backcolour}, commentstyle=\color{codegreen},
  keywordstyle=\color{magenta},
  numberstyle=\tiny\color{codegray},
  stringstyle=\color{codepurple},
  basicstyle=\ttfamily\scriptsize,
  breakatwhitespace=false,         
  breaklines=true,                 
  captionpos=b,                    
  keepspaces=true,                 
  showspaces=false,                
  showstringspaces=false,
  showtabs=false,                  
  tabsize=2
}


%% file: commands.tex
\newcommand{\algorithmautorefname}{Algorithm}
\renewcommand{\figureautorefname}{Fig.}
\renewcommand{\equationautorefname}{Eq.}
\renewcommand{\sectionautorefname}{Section}
\renewcommand{\subsectionautorefname}{Section}


\newcommand{\casadi}{\texttt{casadi}}
\newcommand{\cusadi}{\texttt{CusADi}}
\newcommand{\etoe}{\textit{End-to-End}}
\newcommand{\singleagent}{\textit{Single-Agent}}
\newcommand{\multiagent}{\textit{Multi-Agent}}

\newcommand{\twist}{\boldsymbol{\mathcal{V}}}   
\newcommand{\palmTwist}{\boldsymbol{\mathcal{V}}_P}   
\newcommand{\linVel}{\boldsymbol{v}}            
\newcommand{\linVelX}{v_x}                      
\newcommand{\linVelY}{v_y}                      
\newcommand{\linVelZ}{v_z}                      
\newcommand{\angVel}{\boldsymbol{\omega}}       
\newcommand{\angVelX}{\omega_x}                 
\newcommand{\angVelY}{\omega_y}                 
\newcommand{\angVelZ}{\omega_z}                 

\newcommand{\genPos}{\mathbf{q}}                            
\newcommand{\genVel}{\dot{\mathbf{q}}}                      
\newcommand{\genVelDes}{\dot{\mathbf{q}}_{\text{des}}}      
\newcommand{\genAcc}{\ddot{\mathbf{q}}}                     
\newcommand{\jacobian}{\mathbf{J}}                          

\newcommand{\armObs}{\mathcal{O}_{\text{arm}}}                 
\newcommand{\armJointPos}{\mathbf{q}_{\text{arm}}}             
\newcommand{\armJointVel}{\dot{\mathbf{q}}_{\text{arm}}}       
\newcommand{\prevArmAction}{\mathbf{a}_{\text{arm}}^{t-1}}     
\newcommand{\currArmAction}{\mathbf{a}_{\text{arm}}^{t}}       
\newcommand{\handObs}{\mathcal{O}_{\text{hand}}}               
\newcommand{\handJointPos}{\mathbf{q}_{\text{hand}}}           
\newcommand{\handJointVel}{\dot{\mathbf{q}}_{\text{hand}}}     
\newcommand{\prevHandAction}{\mathbf{a}_{\text{hand}}^{t-1}}   
\newcommand{\currHandAction}{\mathbf{a}_{\text{hand}}^{t}}     
\newcommand{\objState}{\mathbf{s}_{\text{obj}}}                
\newcommand{\objPos}{\mathbf{p}_{\text{obj}}}                  
\newcommand{\objOri}{\mathbf{R}_{\text{obj}}}                  
\newcommand{\objDim}{\mathbf{d}_{\text{obj}}}                  

\newcommand{\armAction}{\mathcal{A}_{\text{arm}}}                       
\newcommand{\armJointPosDes}{\Delta\hat{\boldsymbol{q}}_{\text{arm}}}   
\newcommand{\handAction}{\mathcal{A}_{\text{hand}}}                     
\newcommand{\handJointPosDes}{\Delta\hat{\boldsymbol{q}}_{\text{hand}}} 
\newcommand{\refJointPos}{\boldsymbol{q}^\text{ref}}                    

\newcommand{\zeroVec}{\textbf{0}}       
\newcommand{\world}{\mathcal{W}}             
\newcommand{\palm}{\mathcal{P}}             

\newcommand{\policy}{\boldsymbol{\pi}}      
\newcommand{\gradient}{\hat{g}}             
\newcommand{\advantage}{\hat{A}}            


\newcommand{\comment}[1]{{\color{blue}Comment: #1}}
\newcommand{\hojae}[1]{{\color{cyan}Ho Jae: #1}}
\newcommand{\sehwan}[1]{{\color{orange}Se Hwan: #1}}
\newcommand{\yhlee}[1]{{\color{blue}Yonghyeon: #1}}
\newcommand{\mitmanipulator}{a manipulator platform}%

\newcommand{\todo}[1]{{\color{orange}TODO: #1}}
\newcommand{\fixme}[1]{{\color{red}FIXME: #1}}
\newcommand{\needref}{{\color{blue}[REF]}}

\newcommand{\ie}{i\/.\/e\/.,\/~}%
\newcommand{\eg}{e\/.\/g\/.,\/~}%
\newcommand{\cf}{cf\/.\/~}%

%% file: Sections/00_abstract.tex
\begin{abstract}
    
In this work, we propose a hybrid hierarchical control framework for reactive dexterous grasping that explicitly decouples high-level spatial intent from low-level joint execution. 
We introduce a multi-agent reinforcement learning architecture, specialized into distinct arm and hand agents, that acts as a high-level planner by generating desired task-space velocity commands.
These commands are then processed by a GPU-parallelized quadratic programming controller, which translates them into feasible joint velocities while strictly enforcing kinematic limits and collision avoidance. 
This structural isolation not only accelerates training convergence but also strictly enforces hardware safety. 
Furthermore, the architecture unlocks zero-shot steerability, allowing system operators to dynamically adjust safety margins and avoid dynamic obstacles without retraining the policy. 
We extensively validate the proposed framework through a rigorous simulation-to-reality pipeline. 
Real-world hardware experiments on a 7-DoF arm equipped with a 20-DoF anthropomorphic hand demonstrate highly robust zero-shot transferability for dexterous grasping to a diverse set of unseen objects, highlighting the system's ability to reactively recover from unexpected physical disturbances in unstructured environments.

\end{abstract}

%% file: Sections/01_intro_v2.tex
\section{INTRODUCTION}

Reactive dexterous grasping with multi-degree-of-freedom (DoF) hands remains a long-standing challenge in robotics. 
Its contact-rich nature gives rise to hybrid dynamics, in which contact modes change discontinuously and frictional interactions are complex. 
Unlike parallel grippers, where grasping can often be reduced to forming an antipodal grasp and applying sufficient squeezing along a dominant DoF, dexterous multi-finger grasping requires coordinated control of contact placement and force distribution under the constraints of high-dimensional hand kinematics and object geometry. 
As a result, successful grasping depends on continuous reactive adaptation of both motion and contact forces to evolving contact conditions.

A common approach to grasp execution is to decompose it into three stages: (i) grasp pose synthesis based on criteria such as force closure~\cite{lynch2017modern}, (ii) collision-free motion planning, and (iii) trajectory-tracking control~\cite{zurbrugg2025graspqp, chen2025dexonomy}. 
While this sequential pipeline can be effective in static, structured environments, it is less suited for dynamic, unstructured settings that require online adaptation. 
In particular, grasp synthesis and motion planning introduce nontrivial computational overhead, while standard trajectory tracking offers limited reactivity to disturbances and environmental changes.

Recent contact-implicit model predictive control (MPC) formulations offer a principled physics-based framework for reactive manipulation and grasping with relatively few assumptions~\cite{suh2025dexterous,escande2013planning,pang2023global,posa2014direct}. 
They do not require, in principle, a precomputed grasp pose and allow online grasp synthesis and reactive control. 
In practice, however, these methods often depend on accurate object geometry and reliable contact-state estimation, while trading physical fidelity for tractability through simplified geometric models or relaxed contact dynamics. 
Furthermore, many are derived under quasi-static or quasi-dynamic assumptions, which limit their applicability to dynamic grasping tasks.

A promising recent trend is the use of reinforcement learning (RL) to train reactive grasping policies at scale in simulation~\cite{singh2024dextrah,zhang2025robustdexgrasp}. 
This eliminates the need for explicit object geometry, contact-state estimation, or precomputed grasp poses at inference time. 
However, RL-based approaches still face three key limitations. 
First, training is computationally expensive and scales poorly with the diversity of objects and interaction scenarios. 
Second, the learned policy may behave unpredictably, while safety is typically enforced only indirectly through reward shaping and therefore lacks formal guarantees. 
Third, the resulting policy is often difficult to steer or adapt reliably when the environment changes, for example, in the presence of previously unseen obstacles.

We argue that a key source of these challenges is the excessive learning burden placed on a single policy. 
In many RL-based grasping systems, one policy is expected to simultaneously solve (i) high-level task reasoning, such as how to approach and grasp an object in task space; (ii) low-level kinematic coordination, such as controlling many arm and finger joints to realize the desired task-space behavior; and (iii) safety and constraint satisfaction, including collision avoidance and physical limits. 
These objectives are typically combined through multiple reward terms, which are difficult to tune and often conflict with one another. 
Consequently, the learned policy may struggle to balance them effectively, resulting in substantial reward-engineering effort, slow training, and suboptimal performance.

To address these limitations, we propose a hybrid hierarchical control framework for reactive dexterous grasping that decouples task-level planning from joint-level execution by construction.
The proposed framework improves data efficiency, enables explicit safety and constraint handling, and preserves steerability through online modulation of motion direction and speed.
Our key principle is simple: \textit{let physics handle what is tractable by design, and use RL only where modeling becomes difficult}.

Specifically, the proposed architecture consists of two layers: a high-level planner outputs task-space velocity commands, while a low-level controller solves a quadratic program (QP) to map these commands to joint velocity commands subject to constraints such as collision avoidance, joint velocity limits, and joint position limits. 
In this way, learning is confined to high-level task-space reasoning, while low-level kinematic coordination and constraint handling are handled explicitly. 
By decoupling task-space decision-making from low-level execution, the framework reduces the burden on RL, improves data efficiency, and enables explicit safety-aware control.

The third issue, steerability, is addressed through two complementary mechanisms. 
First, the explicit task-space velocity commands generated by the high-level policy can be modulated online to avoid previously unseen obstacles using existing techniques such as artificial potential fields (APF) or velocity-field-based control~\cite{khatib1986real, huber2023avoidance, lee2025behavior}.
Second, the degree of conservativeness in execution, including motion speed and collision-avoidance margin, can be adjusted at the lower layer through the constraints encoded in the QP. 
Together, these design choices enable the proposed hybrid hierarchical control architecture to address all three challenges simultaneously.

In addition, we decompose the high-level policy into two specialized agents, an arm policy for global palm transport and a hand policy for fine-grained fingertip manipulation, and jointly train them within a multi-agent RL framework. 
This functional decomposition is inspired by literature from human motor control, in which global reaching and local grasp formation are coordinated through distinct proximal and distal mechanisms \cite{graziano2007mapping,santello1998postural,jeannerod1984timing,castiello2000reach}. 
Such a separation enables simultaneous global transport and local adjustment during grasping. 
From the RL perspective, it also reduces conflicting reward signals and gradient interference between these roles~\cite{lee2025learning, yu2020gradient, schaul2019ray}. 
As a result, the two sub-policies converge more reliably and achieve better overall performance than a single monolithic policy trained to learn both behaviors at once.

In the proposed framework, a key technical challenge is to integrate the low-level QP controller into the RL pipeline. 
Crucially, its constraints must be enforced during training rather than imposed only at deployment, so that the high-level policy learns to operate through the low-level controller. 
To this end, inspired by~\cite{jeon2024cusadi}, we develop and open-source a GPU-parallelized QP solver and embed it directly into the RL training loop. 
Although developed here for reactive grasping, the proposed control framework, RL architecture, and solver are broadly applicable to other manipulation tasks.

Finally, we extensively validate the proposed framework on a 7-DoF arm equipped with a 20-DoF five-finger (5F) anthropomorphic hand (ROBOTIS HX5-D20). 
Across comprehensive simulation and real-world experiments, we show that, using only object position tracking and coarse size information, and without requiring explicit object 3D meshes or precomputed grasp poses, the proposed method achieves robust grasping across a diverse set of objects and grasp configurations while strictly satisfying kinematic constraints. 
We further demonstrate in simulation that the same framework extends to an 8-DoF two-finger (2F) gripper.

\begin{figure}[!t]
    \centering
    \includegraphics[width=0.9\linewidth]{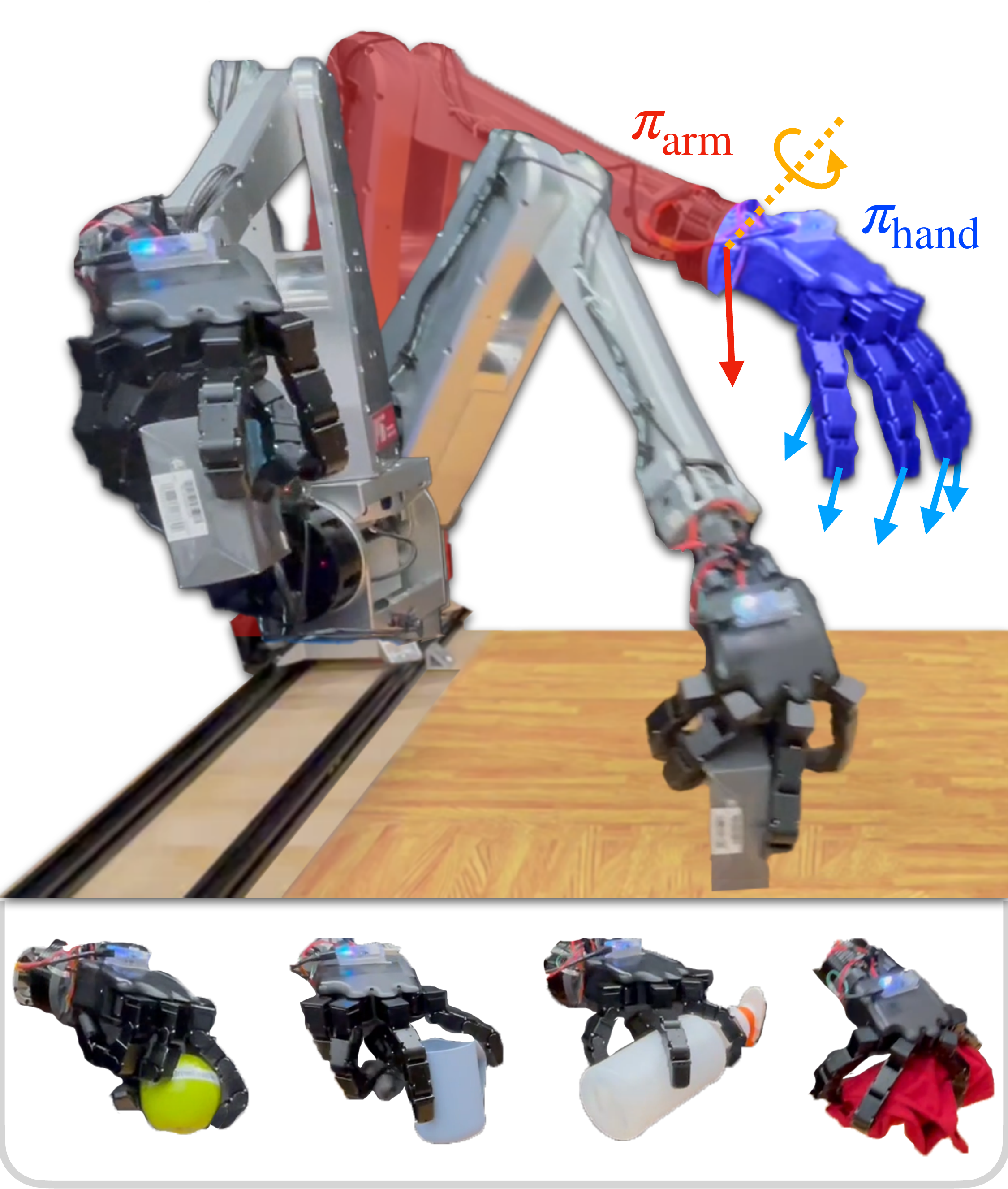}
    \caption{
    Overview of the proposed multi-agent RL framework with hardware validation. 
    Our manipulator platform, equipped with a five-finger anthropomorphic hand, successfully grasps and lifts a target object. 
    The background overlay illustrates the policy decomposition. 
    The arm agent $\pi_{\text{arm}}$ (red) commands the palm twist (red and orange arrows), while the hand agent $\pi_{\text{hand}}$ (blue) independently commands the local fingertip linear velocities (light blue arrows).
    The bottom row highlights successful grasps on arbitrary object shapes and deformables.
    }
    \label{fig:main}
    \vspace{-2mm}
\end{figure}

We summarize our contributions as follows:
\begin{itemize}
\item We propose a hybrid hierarchical control architecture that combines RL with an explicit physics-based controller, enabling more data-efficient learning, explicit constraint handling, and steerable behavior.

\item We show that decomposing the RL policy into palm-transport and finger-manipulation components, and training them within a multi-agent RL framework, yields higher overall performance than a monolithic policy.

\item We develop and open-source\footnote{The source code will be released upon publication.} a GPU-parallelized QP solver that is integrated directly into the RL training pipeline and is broadly applicable to arm--hand manipulation tasks beyond grasping.

\item Through extensive simulation and hardware experiments, we demonstrate that the proposed framework achieves robust grasping across a variety of objects without explicit object models or precomputed grasp poses\footnote{\href{https://youtu.be/2EWOQwZcX9E}{Supplementary video}: https://youtu.be/2EWOQwZcX9E}.
\end{itemize}

%% file: Sections/02_related_works_v2.tex
\section{Related Works}
\label{sec:related_works}

\subsection{Model-Based Methods for Reactive Control and Grasping}

Reactive control refers to the ability of a system to respond and adapt compliantly in real time to environmental changes and external disturbances. 
In contrast to simply tracking time-parametrized motions planned offline, reactive control relies on feedback-based motion generation and compliant control.

Early approaches to collision-free motion generation with closed-loop reactivity were based on APF~\cite{khatib1986real, rimon1990exact, lacevic2013safety, paternain2017navigation}. 
More recent methods have adopted dynamical systems (DS) formulations~\cite{goncalves2010vector, khansari2012dynamical, huber2022avoiding, billard2022learning,lee2025behavior}, in which local modulation of a nominal motion field enables efficient obstacle avoidance through analytical constructions. 
Several recent variants further provide formal convergence guarantees for specific classes of obstacles, such as convex objects, star-shaped geometries, and tree-of-stars environments~\cite{huber2023avoidance}.

However, these methods are generally tailored to relatively simple obstacle geometries, do not readily accommodate general non-convex constraints, and typically assume a prespecified goal configuration~\cite{koptev2024reactive,lee2025hierarchical}. 
Consequently, they are not well-suited for dexterous grasping of arbitrarily shaped objects without an explicit goal pose. 
A recent work addressed part of this challenge by proposing a hierarchical framework for reactive grasping of highly concave objects, such as wine glasses~\cite{lee2025hierarchical}. 
While the method provides a practical solution for the 2F gripper, it still relies on predefined grasp points, whose specification becomes nontrivial for multi-fingered hands.

Another attractive alternative that avoids specifying a precomputed grasp or goal pose is contact-implicit MPC, where contact modes and motions can, in principle, be optimized jointly under task-level objectives~\cite{kim2022contact,posa2014direct,suh2025dexterous,pang2023global}. 
In practice, however, these methods typically require accurate object geometry and reliable contact-state estimation, and often depend on relaxations of contact dynamics and geometric constraints for computational tractability. 
Furthermore, with sparse objective functions, the optimization is generally still too expensive for real-time control and is therefore more suitable for offline trajectory generation. 
Their reliance on quasi-static or simplified quasi-dynamic assumptions also limits their effectiveness for dynamic dexterous grasping.

To overcome these limitations, we use RL to learn reactive grasping behaviors offline and amortize them into a policy that can be executed online with a single neural network forward pass, avoiding precomputed grasp poses and costly real-time global trajectory optimization. 
Importantly, unlike pure RL methods, our approach incorporates model-based priors, as described in the next section.

\subsection{RL for Dexterous Grasping}
RL has emerged as a dominant paradigm for dexterous manipulation, offering an alternative to explicit modeling and reasoning over contact dynamics~\cite{yu2022dexterous}. 
Foundational works have demonstrated the remarkable effectiveness of model-free RL in learning highly intricate behaviors, including continuous in-hand object reorientation and complex tool-use tasks~\cite{andrychowicz2020learning,rajeswaran2017learning}.

The same paradigm has also been explored in dexterous grasping, with recent works showing that RL can learn robust grasping policies across diverse objects and grasp configurations~\cite{zhang2025robustdexgrasp, wan2023unidexgrasp++, agarwal2023dexterous}. 
Such approaches enable reactive grasping without precise object geometry, predefined grasp poses, or computationally expensive real-time global motion planning.

These approaches, however, rely on monolithic end-to-end learning architectures, in which a single neural network maps high-dimensional observations directly to low-level joint commands, such as target joint positions or actuator torques, or to a PCA-reduced action space. 
This paradigm forces the policy to simultaneously handle several distinct objectives, including high-level task planning, low-level kinematic execution, and constraint or safety satisfaction \cite{levine2016end}. 
As a result, the learning problem becomes overly demanding, often leading to slow convergence and suboptimal performance, as we later demonstrate in our experiments.
Moreover, the large exploration burden typically requires substantial reward shaping or human demonstrations to induce effective behavior \cite{mandlekar2021matters}, and the resulting policies often produce jerky and physically unstable motions when deployed on real hardware.

To mitigate the limitations of monolithic architectures, recent work has increasingly explored hierarchical reinforcement learning (HRL), in which high-level and low-level policies operate jointly. 
In such frameworks, the high-level policy outputs operational-space poses or sub-goals and can therefore focus on spatial reasoning, while the low-level policy handles execution \cite{martin2019variable}. 
For multi-fingered dexterous hands, hierarchical formulations have been shown to reduce the dimensionality of the exploration space, enabling agents to more efficiently learn complex finger coordination and tool manipulation skills \cite{sancaktar2022curious, gupta2019relay}. 
Despite these architectural improvements, many existing frameworks still rely on learned policies at both levels. 
Delegating low-level execution to learned policies introduces architectural rigidity, as any modification to operational parameters or constraints requires computationally expensive retraining.
In contrast, our approach replaces the low-level learned controller with a physics-based module, thereby reducing the burden on learning and improving efficiency, interpretability, and constraint handling.

A closely related line of work uses model-based priors for low-level execution in reactive grasping~\cite{lum2024dextrah, singh2024dextrah}. 
These methods use low-dimensional actions to parameterize DS motion primitives, specifically geometric fabrics, and handle collision avoidance and joint limits through soft repulsive terms. 
In contrast, we enforce constraints explicitly through a QP layer, often requiring little or no tuning.
Implementation-wise, they rely on PCA-based hand action reduction, which tends to limit grasp diversity, whereas we preserve a richer hand action space and obtain more diverse grasps.


%% file: Sections/03_preliminaries_v2.tex
\section{Preliminaries}

In this section, we briefly introduce two key preliminaries of our framework. 
First, we introduce the multi-agent RL framework employed to train two specialized grasping agents: an arm agent for global hand transport and a hand agent for local fine-grained finger control.
Second, we describe the parallel GPU implementation of the constrained QP, which is critical for efficient training. 

\begin{figure*}[tb]
    \centering
    \includegraphics[width=0.98\textwidth]{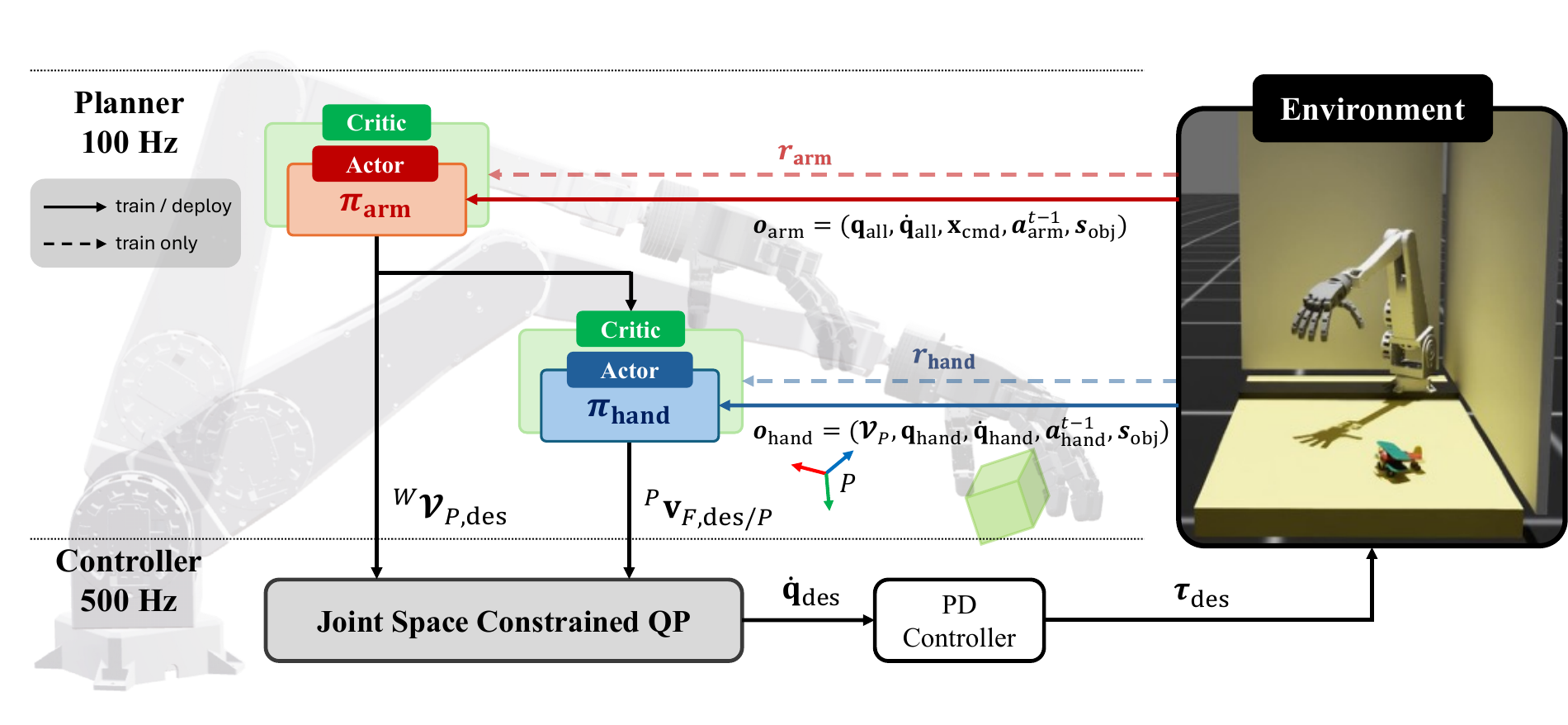}
    \caption{
    Overview of the hybrid hierarchical control framework. 
    Our architecture consists of a high-level RL planner operating at 100 Hz and a low-level QP controller running at 500 Hz. 
    The planner is decomposed into two specialized agents: (i) an arm agent that generates a global transport strategy via a desired palm twist ${}^{W}\twist_{P, \text{des}}$ in the world frame, and (ii) a hand agent that generates local grasping strategy via desired fingertip linear velocities ${}^{P}\mathbf{v}_{F, \text{des}/P}$ relative to the palm. 
    This decomposition enables a unified global-to-local manipulation strategy. 
    These task-space velocities are processed by a joint-space constrained QP controller to compute optimal joint velocities $\dot{\mathbf{q}}_{\text{des}}$ while strictly enforcing physical safety and hardware constraints. 
    This hierarchy allows the policies to focus on learning high-level strategy by delegating physical feasibility and complex differential kinematics to the low-level controller.
    }
    \label{fig:overview}
    \vspace{-2mm}
\end{figure*}

\subsection{Multi-Agent Reinforcement Learning}
\label{sec:marl}


RL~\cite{kober2013reinforcement} is a machine learning paradigm where an agent learns to make decisions by interacting with an environment to maximize cumulative rewards.
The RL process is formally grounded in a Markov Decision Process (MDP)~\cite{sutton1998reinforcement}, defined by the tuple $\mathcal{M} = \langle S, A, P, P_{0}, R, \gamma \rangle$. 
Here, $S$ and $A$ represent the state and action space, $P(s_{t+1}|s_{t},a_{t})$ denotes the transition probability from state $s_{t}$ to state $s_{t+1}$ given action $a_{t}$, $P_{0}$ is the initial state distribution, $R$ is the reward function, and $\gamma \in [0,1)$ is the discount factor for future rewards. 
The goal of RL is then to determine an optimal policy $\pi^{*}(a|s)$ that maximizes the expected cumulative reward over the discounted infinite horizon trajectory:
\begin{equation}
\pi^{*}(a|s) = \arg\max_{\pi} \mathbb{E} \left[ \sum_{t=0}^{\infty} \gamma^{t} R(s_{t}, a_{t}) \right]
\end{equation}

Multi-agent reinforcement learning (MARL) is a powerful paradigm widely employed in complex environments that require distributed decision-making and coordinated behaviors~\cite{gronauer2022multi}. 
In the context of high-dimensional robotics, MARL effectively mitigates the scaling limitations of standard RL by separating control roles and distributing the task responsibilities across actors~\cite{lee2025learning}.
In a general MARL setting, each agent maintains a separate actor policy, utilizes a separate value function, and optimizes a distinct reward function based on its local observations. 
However, this independent learning introduces a critical challenge: because multiple agents are updating their policies simultaneously, the environmental dynamics become non-stationary from the perspective of any single agent.
A widely established solution to this non-stationarity is the Centralized Training with Decentralized Execution (CTDE) paradigm \cite{lowe2017multi, foerster2018counterfactual}. 
CTDE stabilizes learning by granting each agent's centralized critic access to global information during the offline training phase, encompassing the aggregated observations and actions of all independently acting agents in the environment. 
This global visibility ensures a stationary environment for accurate gradient updates. During physical deployment, each agent executes its policy in a purely decentralized manner, relying solely on its individual local observation.


\subsection{Constrained QP and Parallelization in GPU}
\label{sec:qp_parallelization}

QP is the process of solving mathematical optimization problems with quadratic costs and linear constraints. 
It is widely used in robotic control because many control problems can be solved by forming local linear approximations of the kinematics and dynamics.
In this section, we introduce the standard constrained QP formulation and describe our GPU-parallelized implementation, which enables seamless integration into large-scale RL pipelines. 

The general form of the QP is:
\begin{equation}
\label{eq:qp_formulation}
\begin{aligned}
\min_{\mathbf{x}} \quad & \frac{1}{2} \mathbf{x}^\top \mathbf{H} \mathbf{x} + \mathbf{g}^\top \mathbf{x} \\
\text{subject to} \quad & \mathbf{A}_{\text{eq}} \mathbf{x} = \mathbf{b}_{\text{eq}} \\
& \mathbf{A}_{\text{ineq}} \mathbf{x} \leq \mathbf{b}_{\text{ineq}}
\end{aligned}
\end{equation}
where $\mathbf{x} \in \mathbb{R}^{n}$ represents the optimization variable, $\mathbf{H} \in \mathbb{S}_{+}^{n}$ denotes the symmetric positive-definite Hessian matrix representing the quadratic cost, $\mathbf{g} \in \mathbb{R}^{n}$ is the gradient vector, and the matrices $\mathbf{A}_{\text{eq}} \in \mathbb{R}^{m \times n}$ and $\mathbf{A}_{\text{ineq}} \in \mathbb{R}^{p \times n}$ define the equality and inequality constraints, respectively. 
The vectors $\mathbf{b}_{\text{eq}} \in \mathbb{R}^{m}$ and $\mathbf{b}_{\text{ineq}} \in \mathbb{R}^{p}$ represent the equality and inequality bounds, respectively, where $m$ and $p$ denote the number of corresponding constraints. 
In the context of robotic manipulation, these parameters typically encode physical constraints such as joint position and velocity limits, torque saturations, collision constraints, or the dynamics model of the robot system.

To parallelize the QP formulation, we adopt the approach described in~\cite{jeon2024cusadi}. 
Specifically, we adopt an Interior Point Method~\cite{boyd2004convex} to handle the inequality constraints. 
In this framework, the original problem (\autoref{eq:qp_formulation}) is approximated by incorporating the inequality constraints into the objective via a logarithmic barrier function. 
The resulting equality-constrained problem is represented as follows:
\begin{equation}
\label{eq:qp_ipm}
\begin{aligned}
\min_{\mathbf{x}} \quad & \frac{1}{2} \mathbf{x}^\top \mathbf{H} \mathbf{x} + \mathbf{g}^\top \mathbf{x} + \mu \cdot \Phi(\mathbf{A}_{\text{ineq}} \mathbf{x} - \mathbf{b}_{\text{ineq}}) \\
\text{subject to} \quad & \mathbf{A}_{\text{eq}} \mathbf{x} = \mathbf{b}_{\text{eq}}
\end{aligned}
\end{equation}
where $\mu > 0$ is a barrier parameter and $\Phi(\cdot)$ denotes the barrier function, typically formulated using quadratic and logarithmic terms to enforce feasibility. 
Given an initial guess for $\mathbf{x}$, we compute a search direction $\Delta \mathbf{x}$ by applying $LDL^\top$ factorization to the KKT system derived from the second-order Taylor approximation near $\mathbf{x}$. 
The optimization variable is then updated via a Newton step:
\begin{equation}
    \mathbf{x} \leftarrow \mathbf{x} + \alpha \Delta \mathbf{x}
\end{equation}
where $\alpha \in (0, 1]$ is a step length determined to ensure the new iterate remains within the feasible interior. 
This approach allows us to formulate the QP solver as a \casadi~function~\cite{andersson2019casadi}, which then can be parallelized on the GPU by \cusadi~\cite{jeon2024cusadi}.
In this work, we leverage this approach to parallelize the low-level joint-space QP controller on the GPU, enabling its seamless integration into the RL training pipeline.

%% file: Sections/04_system_architecture.tex
\section{System Architecture}
\label{sec:system_architecture}



In this section, we outline the overall system architecture designed to achieve reactive dexterous grasping for arbitrarily shaped objects using a multi-fingered hand. 
As described in \autoref{fig:overview}, our architecture is bifurcated into two main components: \textit{a high-level RL planner} and \textit{a low-level constrained QP controller}.

The RL planner performs high-level reasoning and generates the desired palm twist -- i.e., angular and linear velocities --  and fingertip linear velocities, given the robot and target object state. 
The constrained QP controller then translates these high-level velocity commands into joint velocity commands while ensuring that the robot adheres to physical and safety constraints. 
A proportional-derivative (PD) controller subsequently transforms these joint-velocity commands into torque commands for the motors.
This architectural separation allows the RL process to focus exclusively on discovering complex task-level strategies, which are often the most critical for success yet remain the most challenging to model manually (e.g., grasp poses).
Reflecting their distinct roles, the high-level planner operates at 100 Hz, where a slower rate suffices for strategic decision-making, while the low-level QP controller runs at 500 Hz to enforce constraints and handle safety-critical events in real time.

This architecture is designed for generalizability, allowing it to be applied to any fixed-base manipulator comprised of an arm and a multi-fingered end-effector. 
Our manipulator platform (\autoref{fig:manipulator}) is adapted from the system introduced in \cite{saloutos2022towards}. 
Our platform supports either a 2F gripper (8 DoF) or a 5F anthropomorphic hand (20 DoF, with 4 DoF per finger), both of which interface with a shared 7-DoF arm assembly.


The training pipeline is implemented in the IsaacLab simulation environment \cite{mittal2025isaac}, which supports GPU parallelization for RL. 
The architectural details of each component are presented in the following sections: \autoref{sec:rl_planning} describes the task space RL planner with a multi-agent learning framework for the arm and hand, while \autoref{sec:qp_control} presents the low-level joint space constrained QP control formulation and its GPU parallelization for integration with RL.

%% file: Sections/05_rl_planning.tex
\section{Multi-Agent RL Task Space Planning}
\label{sec:rl_planning}

\begin{figure}[t]
    \centering
    \includegraphics[width=0.98\linewidth]{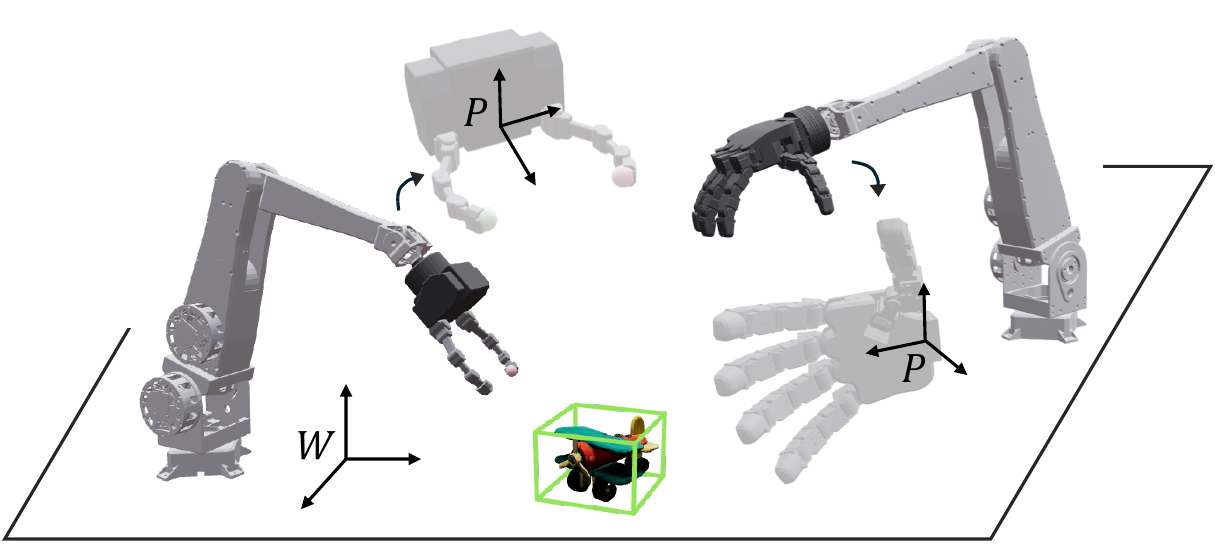}
    \caption{
    Illustration of the fixed-based manipulation platforms and coordinate frames. 
    \textit{Left:} Two-fingered gripper.
    \textit{Right:} Five-fingered anthropomorphic hand. 
    Zoomed views show the local palm frame $P$ used for fingertip velocity mapping and the computation of hand-centric local observations.
    The target object is modeled as a minimum volume bounding box (green), enabling the framework to generalize across various arbitrary object shapes.}
    \label{fig:manipulator}
    \vspace{-4mm}
\end{figure}

The first main component of our hierarchical framework is a high-level task-space planner trained via RL.
Given the current robot state and the target object state, the high-level planner generates strategic reaching and manipulation trajectories represented by a desired palm twist and fingertip linear velocities.
Drawing inspiration from the human motor control system, which utilizes distinct neural pathways for reaching and grasping \cite{graziano2007mapping, santello1998postural, jeannerod1984timing, castiello2000reach}, we adopt a multi-agent RL architecture. 
Specifically, we decompose the planner into two specialized agents that coordinate toward a shared manipulation goal:
\begin{itemize}
\item \textbf{Arm Agent:} Plans the motion of the palm (the base of the hand), focusing on transporting the end-effector toward the target object.
The action is represented by a desired palm twist (i.e., linear and angular velocities) described in the world frame $\{W\}$.
\item \textbf{Hand Agent:} Plans the motion of the fingertips, directing the dexterous manipulation and securing of the object.
The action is represented by desired fingertip linear velocities described in the palm frame $\{P\}$. 
\end{itemize}

Under this paradigm, the arm and hand agents concurrently output a desired palm twist and fingertip linear velocities, respectively, forming a unified global-to-local manipulation strategy. 
This decoupling allows each agent to have its own tailored observation space and reward function, while preventing policy gradient updates of one agent from interfering with those of the other, as discussed in Section~\ref{sec:marl}. 
The rest of this section details the RL planner, including the observation spaces, action spaces, and reward structures for both agents.



\subsection{Arm agent}
\label{sec:arm_agent}

The primary role of the arm is to transport the hand to an optimal pre-grasp pose around the object to facilitate the hand's grasping task. 
The observation space of the arm policy, denoted as $\armObs$, includes proprioceptive robot state, the target object state, and the palm's lift pose command. 

Specifically, the proprioceptive state consists of the full joint positions $\genPos$ and velocities $\genVel$ of the arm and hand, alongside the previous arm action $\prevArmAction$.
The policy also observes the target object state $\objState$, represented by its Minimum Volume Bounding Box (MVBB). This bounding box captures the spatial position $\objPos$, orientation $\objOri$, and dimensions $\objDim$ of the object. 
To accommodate the geometric variability of target objects, the MVBB provides a standardized, low-dimensional representation of spatial extent and 6D pose. 
This abstraction ensures a consistent observation space, enabling the arm agent to effectively generalize its approach across diverse object shapes.
Lastly, the observation space includes the predefined target lift pose command $\mathbf{x}_{\text{cmd}} = [\mathbf{p}_{\text{cmd}}^\top, \mathbf{Q}_{\text{cmd}}^\top]^\top$. This command defines the desired 6D pose that the palm must track after a successful grasp, represented as a concatenated vector of the 3D Cartesian position $\mathbf{p}_{\text{cmd}} \in \mathbb{R}^{3} $ and the 4D quaternion orientation $\mathbf{Q}_{\text{cmd}} \in \mathbb{H}$.
A detailed breakdown of the arm policy's observation space for both platforms is provided in \autoref{tab:arm_obs} in the Appendix.

To enable centralized training, the critic of the arm agent receives an expanded global observation. 
This encompasses the entire actor observation alongside the previous hand action $\prevHandAction$, providing the critic with complete global information of the system state.
The action space $\armAction\in\mathbb{R}^{6}$ corresponds to the desired palm twist ${}^{W}\boldsymbol{\mathcal{V}}_{P, \text{des}} = [{}^{W}\boldsymbol{\omega}_{P, \text{des}}^\top, {}^{W}\mathbf{v}_{P, \text{des}}^\top]^\top$, where ${}^{W}\boldsymbol{\omega}_{P, \text{des}} \in \mathbb{R}^{3}$ and ${}^{W}\mathbf{v}_{P, \text{des}} \in \mathbb{R}^{3}$ denote the desired angular and linear velocities of the palm frame $\{P\}$ with respect to the world frame $\{W\}$.

We design the arm policy's reward function to encourage a strategic reaching behavior that effectively positions the palm for successful grasping by the hand agent.
To guide this behavior, we formulate the following four major rewards:

\textit{1) Palm-to-Object Euclidean Distance:}
During the initial reaching stage, we encourage the arm to minimize the distance between the palm surface and the target object, which is a fundamental prerequisite for successful grasping. However, to prevent the palm from moving excessively close and to preserve the kinematic discretion required to learn optimal grasp poses, we guide the palm to stay within a designated ``Golden Zone''. This boundary is defined by a geometrically informed threshold $\epsilon_d = R_{\text{obj}} + 0.5 L_{\text{finger}}$, calculated as the circumradius $R_{\text{obj}}$ of the object's MVBB plus half of the finger length $L_{\text{finger}}$.
First, defining the palm-to-object relative vector $\mathbf{p}_{PO}$ and its Euclidean distance $d_{PO}$:
\begin{eqnarray}
\mathbf{p}_{PO} &=& \mathbf{p}_{\text{obj}} - \mathbf{p}_{\text{palm}} \\
d_{PO} &=& \|\mathbf{p}_{PO}\|_2 
\end{eqnarray}
we formulate the proximity reward $r_{\text{dist}}$ as follows:
\begin{eqnarray}
r_{\text{dist}} &=& \exp \left( -\frac{\max(0, d_{PO} - \epsilon_d)^2}{\sigma_{\text{dist}}} \right)
\end{eqnarray}
This formulation allows the arm agent to transport the hand sufficiently close to the target, while preserving the freedom to refine its own last-centimeter positioning during active coordination with the hand agent.

\textit{2) Palm Orientation Alignment to Palm-to-Object Vector:}
In addition to transporting the hand close to the target, the arm agent must also achieve a proper pre-grasp orientation. Because the hand naturally secures an object by pulling it toward the palm, we explicitly encourage the palm surface's normal axis to face the object.

Since the palm surface's normal vector $\mathbf{z}_P$ and the normalized palm-to-object vector $\hat{\mathbf{p}}_{PO}$ are both unit vectors constrained to the surface of a unit sphere, we measure the shortest path between them using the geodesic angular distance $\theta$:
\begin{eqnarray}
\hat{\mathbf{p}}_{PO} &=& \frac{\mathbf{p}_{PO}}{\|\mathbf{p}_{PO}\|_2} \\
\theta &=& \arccos\left(\mathbf{z}_P^\top \hat{\mathbf{p}}_{PO}\right)
\end{eqnarray}
To prioritize spatial translation during the early stages of the trajectory, the orientation alignment reward $r_{\text{align}}$ is multiplicatively gated by the current proximity reward $r_{\text{dist}}$. 
Using the computed angular distance $\theta$, we formulate this reward as follows:
\begin{eqnarray}
r_{\text{align}} &=& r_{\text{dist}} \cdot \exp \left( -\frac{\max(0, \theta - \epsilon_\theta)^2}{\sigma_{\text{align}}} \right)
\end{eqnarray}
Here, an angular deadband $\epsilon_\theta$ of $30^\circ$ is introduced to provide the policy with sufficient functional freedom. 
This prevents over-constraining the wrist kinematics, allowing the agent to successfully execute diverse strategies such as off-center and side approaches.

\textit{3) Stable Grasp:}
Establishing contact and securing a firm grasp is not the sole responsibility of the hand agent; rather, it is a highly collaborative effort. The arm must navigate the hand into a proper pre-grasp pose while the fingers dynamically articulate to establish secure contact. To foster this multi-agent coordination, we define a shared grasp indicator $\mathbb{I}_{\text{grasp}}$ that requires both sufficient palm proximity and active object engagement:
\begin{eqnarray}
\mathbb{I}_{\text{prox}} &=& \begin{cases} 1, & d_{PO} < \epsilon_d \\ 0, & \text{otherwise} \end{cases} \\
\mathbb{I}_{\text{contact}} &=& \begin{cases} 1, & N_{\text{contacts}} > 2 \\ 0, & \text{otherwise} \end{cases} \\
\mathbb{I}_{\text{grasp}} &=& \mathbb{I}_{\text{prox}} \cdot \mathbb{I}_{\text{contact}} 
\end{eqnarray}
According to the classical theory of form closure \cite{lynch2017modern}, a spatial 3D body requires a minimum of seven frictionless point contacts to achieve a stable grasp. To encourage the emergence of highly robust, enveloping grasps that satisfy form closure, we reward the policy based on the number of valid contact points, capped at a maximum of seven. The final grasp reward $r_{\text{grasp}}$ is formulated as follows:
\begin{eqnarray}
\label{eq:stable_grasp}
r_{\text{grasp}} &=& \mathbb{I}_{\text{grasp}} \cdot \min(N_{\text{contacts}}, 7)
\end{eqnarray}
By tying the arm's reward directly to the hand's contact success, this formulation incentivizes joint collaboration, ensuring the arm actively maintains an optimal pose while the hand secures the object.

\textit{4) Goal-Conditioned Task-Space Lift:}
Once a stable grasp is achieved, the arm must transport the object to a commanded 6D target pose, defined by a Cartesian position $\mathbf{p}_{\text{cmd}}$ and an orientation quaternion $\mathbf{Q}_{\text{cmd}}$. 
The position error $e_{\text{pos}}$ is measured using the standard Euclidean norm, whereas the orientation error $e_{\text{ori}}$ is computed as the magnitude of the axis-angle representation of the difference quaternion:
\begin{eqnarray}
e_{\text{pos}} &=& \|\mathbf{p}_{\text{palm}} - \mathbf{p}_{\text{cmd}}\|_2 \\
e_{\text{ori}} &=& \|\mathrm{AxisAngle}(\mathbf{Q}_{\text{cmd}} * \mathbf{Q}_{\text{palm}}^{-1})\|_2
\end{eqnarray}
To prioritize spatial translation while still enforcing orientation tracking, the final lift reward blends the positional and rotational tracking terms using a 70/30 weighting scheme.
Furthermore, to guarantee that the arm only attempts to lift once the object is securely held, the final lift reward is multiplicatively gated by the shared grasp indicator $\mathbb{I}_{\text{grasp}}$:
\begin{equation}
r_{\text{lift}} = \mathbb{I}_{\text{grasp}} \cdot \left[ 0.7 \exp \left( -\frac{e_{\text{pos}}^2}{\sigma_{\text{pos}}} \right) + 0.3 \exp \left( -\frac{e_{\text{ori}}^2}{\sigma_{\text{ori}}} \right) \right]
\end{equation}
This formulation enables generalized pick-and-place capabilities, driven entirely by task-space commands.

\subsection{Hand Agent}
\label{sec:hand_agent}

The primary role of the hand agent is to execute the dexterous finger movements required to securely grasp the target object. 
The observation space of the hand policy, denoted as $\mathcal{O}_{\text{hand}}$, includes the proprioceptive hand state, the target object state, and the kinematic state of the palm, which serves as the moving base for the hand. 
Specifically, the proprioceptive state consists of the hand joint positions $\handJointPos$ and velocities $\handJointVel$, alongside the previous hand action $\prevHandAction$. 
The policy also observes the target object state. 
However, to ensure the hand agent learns local manipulation strategies independent of the global arm configuration, this object state is explicitly represented in the local palm surface frame, denoted as ${}^{P}\mathbf{s}_{\text{obj}}$. 
Furthermore, because the hand must continuously adapt to the motion of its base, the observation space additionally includes the current palm twist in the palm frame, ${}^{P}\boldsymbol{\mathcal{V}}_{P} = [{}^{P}\boldsymbol{\omega}_{P}^\top, {}^{P}\mathbf{v}_{P}^\top]^\top$, as well as the current action output of the arm agent $\currArmAction = {}^{W}\twist_{P, \text{des}}$. 
A detailed breakdown of the hand policy's observation space for both platforms is provided in \autoref{tab:hand_obs} in the Appendix.

To enable centralized training, the critic of the hand agent shares the exact same expanded global observation as the arm agent's critic, providing complete global information of the system state. 

Unlike the arm agent, the hand agent operates under strict kinematic constraints. 
Because the fingers possess only 4 DoF (with three parallel joints), the local fingertip Jacobian is rank-deficient for full 6D spatial tracking. 
Consequently, we define the action space $\handAction \in \mathbb{R}^{3|\mathcal{F}|}$ as the concatenated desired 3D linear velocities $\{^{P}\mathbf{v}_{i,\text{des}/P}\}_{i \in \mathcal{F}}$ of all active fingertips.
Specifically, ${}^{P}\mathbf{v}_{i,\text{des}/P}$ denotes the desired linear velocity for fingertip $i$ relative to the palm, expressed in the palm frame $\{P\}$. 
By restricting the action space to Cartesian translation, we intentionally leave the remaining DoF unconstrained in the null space, enabling the kinematically redundant fingers to naturally conform to the object's geometry upon contact. 
The set of active fingertips is defined by $\mathcal{F}$; for a two-finger gripper, $\mathcal{F} = \{\text{left, right}\}$, whereas for a five-finger hand, $\mathcal{F} = \{\text{thumb, index, middle, ring, little}\}$.

To guide the grasping behavior, the hand agent is driven entirely by the sparse, multi-contact grasp reward $r_{\text{grasp}}$ established in \autoref{eq:stable_grasp}. 
A common pitfall in RL for dexterous manipulation is the reliance on heavily engineered dense rewards, such as distance heuristics for individual fingertips, which often trap policies in local minima~\cite{rajeswaran2017learning}. 
By sharing this single form-closure-motivated contact reward with the arm agent, we establish a strictly collaborative objective. 
The arm is tasked with maintaining an optimal pre-grasp pose, allowing the hand agent to discover robust enveloping grasps without restrictive reward shaping.

Finally, to ensure physically viable execution and facilitate sim-to-real transfer, both agents incorporate standard regularization and termination penalties (\autoref{tab:rewards}). 
The smoothness constraints penalize excessively jerky action outputs, while the termination penalty prevents unrealistic numerical artifacts upon any illegal collision, defined as unintended contacts between the robotic arm links and the table.

\begin{table}[h]
\centering
\caption{Regularization Reward}
\label{tab:rewards}
\renewcommand{\arraystretch}{1.2} 
\begin{tabularx}{\columnwidth}{l >{\centering\arraybackslash}X} 
\toprule
\textbf{Reward} & \textbf{Expression} \\
\midrule
Action Smoothness 1st-order & $-\|(\mathbf{a}_t - \mathbf{a}_{t-1})/\Delta t\|_2^2$ \\
Action Smoothness 2nd-order & $-\|(\mathbf{a}_t - 2\mathbf{a}_{t-1} + \mathbf{a}_{t-2})/\Delta t\|_2^2$ \\
Termination & $\begin{cases} -1, & \text{illegal collision} \\ 0, & \text{otherwise} \end{cases}$ \\
\bottomrule
\end{tabularx}
\end{table}

%% file: Sections/06_qp_control.tex
\section{Joint Space Constrained QP Control}
\label{sec:qp_control}

While the multi-agent RL policy successfully generates high-level task-space commands, represented as the desired palm twist ${}^{W}\twist_{P, \text{des}}$ and the fingertip linear velocities ${}^{P}\mathbf{v}_{i,\text{des}/P}$, these spatial commands must be physically realized by the robot's actuators.
Rather than requiring the RL agents to learn complex differential kinematics, collision avoidance, and strict physical joint limits, we delegate these lower-level requirements to a joint-space constrained QP controller. 
This modular separation allows the RL policy to focus purely on discovering optimal manipulation strategies, while the QP controller robustly converts the task-space commands into safe, executable joint velocity commands.

\subsection{Constrained QP Formulation}

We formulate the velocity inverse kinematics as a constrained optimization problem. The objective is to minimize the tracking error between the commanded task-space velocities and the robot's operational space velocities, subject to strict physical constraints:
\begin{align}
\label{eq:constrained_qp}
\genVelDes = \underset{\genVel}{\arg\min}
& \left\| {}^{W}\twist_{P, \text{des}} - {}^{W}\!\jacobian_{P}(\genPos)\,\genVel \right\|^2_2 \nonumber \\
& + \sum_{i \in \mathcal{F}} \left\| {}^{P}\mathbf{v}_{i,\text{des}/P} - {}^{P}\!\jacobian_{v, i/P}(\genPos)\,\genVel \right\|^2_2 \nonumber \\
& + \lambda \left\| \genVel \right\|^2_2
\end{align}
\begin{align}
\text{subject to} \quad
& \Gamma(\genPos) + \frac{d\Gamma(\genPos)}{d\genPos}\,\genVel\,H \ge \varepsilon_{\Gamma}, \label{eq:qp_collision} \\
& \genPos_{\min} \le \genPos + \genVel\,H \le \genPos_{\max}, \label{eq:qp_pos_limit} \\
& \genVel_{\min} \le \genVel \le \genVel_{\max}. \label{eq:qp_vel_limit}
\end{align}

In Eq. \eqref{eq:constrained_qp}, $\genPos \in \mathbb{R}^n$ and $\genVel \in \mathbb{R}^n$ denote the joint position and velocity vectors, respectively. The task-space commands generated by the RL agents, ${}^{W}\boldsymbol{\mathcal{V}}_{P, \text{des}}$ and ${}^{P}\mathbf{v}_{i,\text{des}/P}$, are mapped to the joint space via their corresponding analytical Jacobians: ${}^{W}\!\jacobian_{P}(\genPos)$ denotes the base-to-palm spatial Jacobian, and ${}^{P}\!\jacobian_{v, i/P}(\genPos)$ represents the local translational fingertip Jacobian. 
A detailed derivation of this local fingertip Jacobian is provided in Appendix~\ref{appendix:frame_transformation}.
The term $\lambda$ is a small positive scalar used to regularize the joint velocities.

To ensure physical safety, Eq. \eqref{eq:qp_collision} enforces a linearized distance constraint, where $\Gamma(\genPos)$ computes the shortest distance for robot-environment and self-collision pairs, $H$ is the planning horizon, and $\varepsilon_{\Gamma} > 0$ is a strict safety margin. Furthermore, Eq. \eqref{eq:qp_pos_limit} and \eqref{eq:qp_vel_limit} enforce the hardware-specific joint position and velocity limits. The optimization yields the desired joint velocity $\genVel_{\text{des}}$, which is subsequently converted into torque commands $\boldsymbol{\tau}_{\text{des}}$ via a joint PD controller.

\subsection{GPU Parallelization}

Training the high-level multi-agent RL policy requires simulating thousands of independent environments simultaneously. 
Solving the velocity inverse kinematics sequentially on a CPU for thousands of robots creates a massive computational bottleneck. 
To address this, we parallelize the QP formulation to run entirely on the GPU following the procedures described in \autoref{sec:qp_parallelization}. 
Following the general formulation established in Eq. \eqref{eq:qp_ipm}, we evaluate the scalar constraint margin $h = (\mathbf{A}_{\text{ineq}}\genVel - \mathbf{b}_{\text{ineq}})_j$ for each constraint. We then apply the relaxed barrier function $\Phi(h)$, which combines a quadratic penalty and a logarithmic barrier, adopted from \cite{grandia2023perceptive}:
\begin{equation}
\label{eq:relaxed_barrier}
\Phi(h) =
\begin{cases} 
-\mu \ln(h), & h \ge \delta \\
\frac{\mu}{2} \left( \left( \frac{h - 2\delta}{\delta} \right)^2 - 1 \right) - \mu \ln(\delta), & h < \delta
\end{cases}
\end{equation}
where $\mu > 0$ is the barrier penalty parameter and $\delta > 0$ determines the boundary distance where relaxation begins. 
For strictly feasible states ($h \ge \delta$), it behaves as a standard logarithmic barrier. 
However, if a constraint margin approaches zero or is violated ($h < \delta$), the function transitions into a well-conditioned quadratic penalty.
With this approach, we can solve the velocity inverse kinematics in parallel, allowing seamless integration with RL.

\subsection{Joint PD Control}

Once the GPU-parallelized QP solver computes the safe, desired joint velocities $\genVel_{\text{des}}$, these kinematic commands must be converted into dynamic joint torque commands $\boldsymbol{\tau}_{\text{des}}$ to be executed by the robot. 
We achieve this using a low-level, high-frequency PD controller.

First, the desired joint positions $\genPos_{\text{des}}$ are computed by numerically integrating the QP velocity output over the control time step $\Delta t$:
\begin{equation}
\genPos_{\text{des}} = \genPos + \genVel_{\text{des}} \cdot \Delta t
\end{equation}
Subsequently, the required joint torques are calculated using the standard PD control law:
\begin{equation}
\boldsymbol{\tau}_{\text{des}} = \mathbf{K}_p \cdot (\genPos_{\text{des}} - \genPos) + \mathbf{K}_d \cdot (\genVel_{\text{des}} - \genVel) + \boldsymbol{\tau}_{\text{ff}}
\end{equation}
where $\mathbf{K}_p$ and $\mathbf{K}_d$ are diagonal matrices representing the joint position stiffness and velocity damping gains, respectively, and $\boldsymbol{\tau}_{\text{ff}}$ is the feedforward gravity compensation torque. 
This final torque command is sent directly to the robot's joint actuators, completing the hierarchical pipeline from high-level multi-agent task-space planning to low-level motor actuation.

%% file: Sections/07_results.tex
\section{Experimental Results}
\label{sec:experimental_results}

\begin{figure*}[tb]
    \centering
    \begin{subfigure}[b]{\textwidth}
        \centering
        \includegraphics[width=\textwidth]{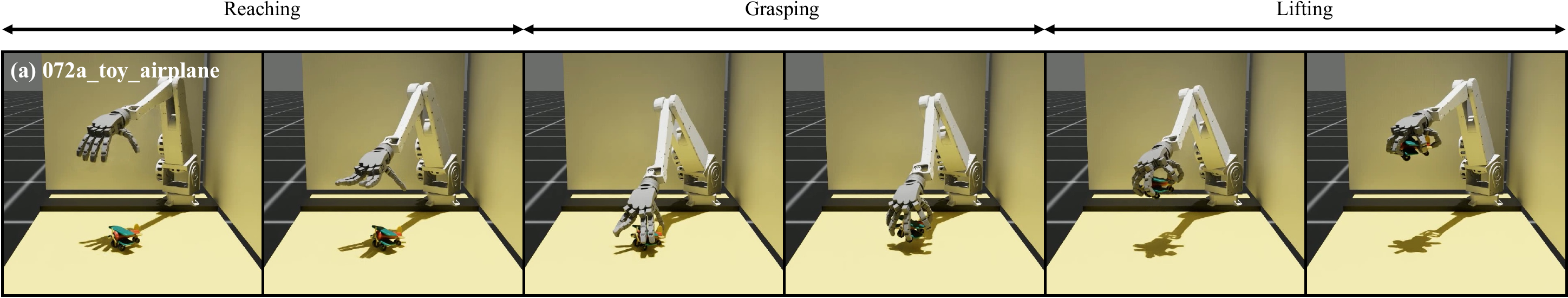}
        \vspace{-3mm} 
    \end{subfigure}
    \begin{subfigure}[b]{\textwidth}
        \centering
        \includegraphics[width=\textwidth]{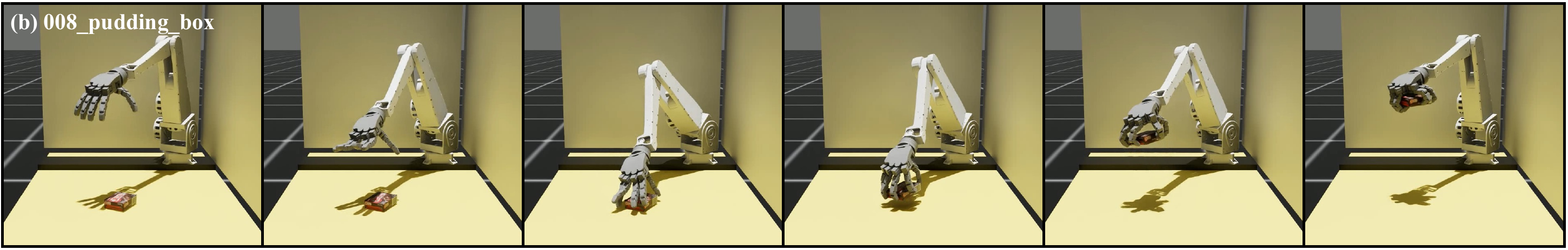}
        \vspace{-3mm} 
    \end{subfigure}
    \begin{subfigure}[b]{\textwidth}
        \centering
        \includegraphics[width=\textwidth]{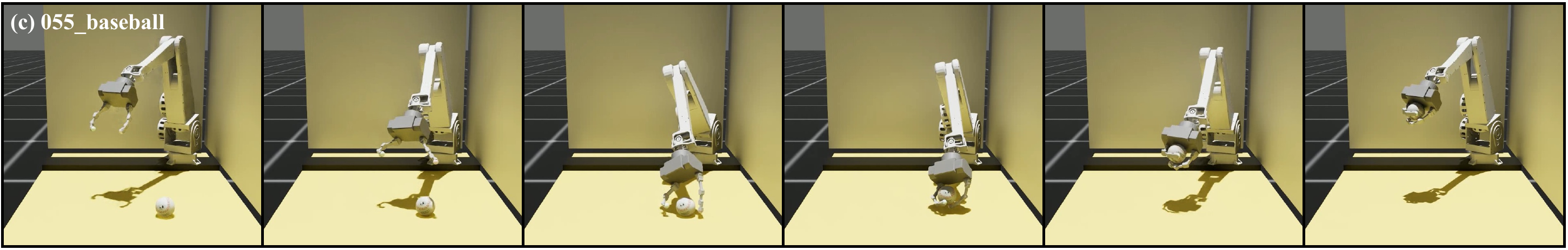}
        \vspace{-3mm} 
    \end{subfigure}
    \begin{subfigure}[b]{\textwidth}
        \centering
        \includegraphics[width=\textwidth]{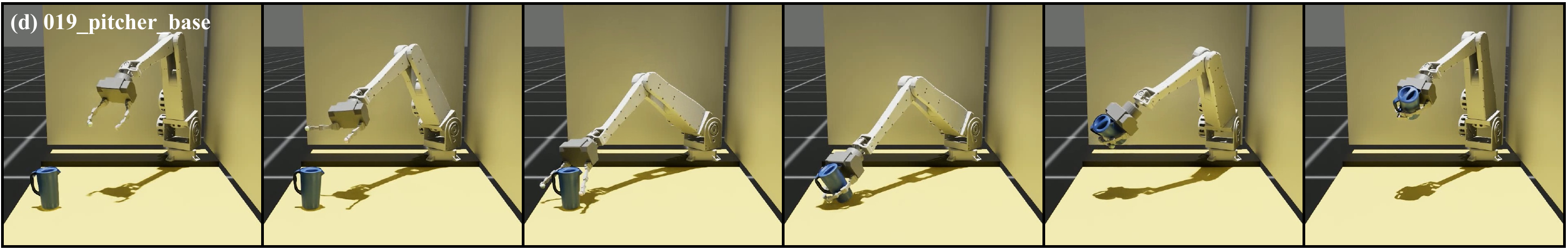}
        \vspace{-3mm} 
    \end{subfigure}
    \caption{
    Simulation results demonstrating the reach-grasp-lift progression of the proposed framework. 
    The top two rows show the 20-DoF 5F hand grasping (a) a toy airplane (ID: 072a) and (b) a pudding box (ID: 008). The bottom two rows show the 8-DoF 2F gripper grasping (c) a baseball (ID: 055) and (d) a pitcher base (ID: 019).
    Our proposed control framework successfully generalizes diverse grasping strategies to various geometries across various end-effector platforms, demonstrating its robustness and adaptability in dexterous grasping tasks.
    }
    \label{fig:sim_results}
\end{figure*}

In this section, we evaluate the proposed hierarchical control framework across large-scale simulations and real-world physical deployments. 
We first detail the training setup in simulation, including the target object scaling curriculum.
Next, we present the overall dexterous grasping performance of our architecture across a diverse set of objects in the simulation environment.
Finally, we demonstrate the zero-shot transfer of our policy to physical hardware, highlighting its robustness during reactive grasping tasks.

\subsection{Simulation Setup}
\label{subsec:simulation_setup}

To robustly train and deploy our policies, we utilize a three-stage Sim-Sim-Real pipeline: policy development and training in IsaacLab~\cite{mittal2025isaac}, deployment validation in MuJoCo~\cite{todorov2012mujoco}, and physical execution on the manipulator hardware platform.
During the development phase, the policies are trained using the Proximal Policy Optimization (PPO) algorithm \cite{schulman2017proximal} across 4,096 parallel environments in IsaacLab. 
To bridge the reality gap, we apply domain randomization to link masses, surface friction, and object properties, as detailed in \autoref{tab:domain_randomization} in the Appendix. 
Once trained, the policy is transferred to a MuJoCo simulation for high-fidelity dynamic validation, utilizing OSQP~\cite{osqp} for the low-level constrained QP control. 
Following this validation phase, the policy is directly deployed to the physical hardware.

To train our policies to generalize across diverse geometric shapes, we utilize objects from the standard YCB dataset~\cite{calli2015ycb}. 
Within the physics simulator, the actual object geometries are modeled using convex decomposition triangular meshes to ensure accurate multi-point contact dynamics. 
However, as established in~\autoref{sec:rl_planning}, the RL agents perceive these objects through an MVBB representation. 
This prevents the policy from overfitting to specific object meshes, encouraging it instead to learn generalized spatial strategies.

Furthermore, to accommodate the specific morphology of our manipulator's hand ($12~\mathrm{cm}$ finger length, $24~\mathrm{cm}$ total hand length), we implemented a standardized scaling protocol for the dataset. 
We constrain the shortest dimension of each object's MVBB to be strictly between $3~\mathrm{cm}$ and $9~\mathrm{cm}$. 
This ensures that all target objects remain within the functional grip aperture of the hand, preventing the learning process from degrading due to physically infeasible grasp attempts, such as precision failures on sub-centimeter objects or insufficient finger wrap on oversized geometries. 
Each YCB object is scaled uniformly to preserve its original aspect ratio while satisfying these spatial constraints. 
Accordingly, the object's mass is proportionally adjusted by the cube of the applied scale factor ($scale^3$) to maintain physically accurate inertial properties. 
To accommodate the maximum kinematic reach of the arm links, the effective workspace of the table where objects are spawned is defined as $0.6~\mathrm{m} \times 0.9~\mathrm{m}$.

\subsection{Simulation Results}
\label{subsec:simulation_results}

The sequential reach-grasp-lift progression of our proposed framework is depicted in \autoref{fig:sim_results}, showing successful executions for both the 5F hand and 2F gripper.
To assess the performance of the trained policies, we establish four quantitative metrics (\autoref{tab:main_results}): success rate, time to success, position error, and orientation error. 
Success is strictly defined as the object being lifted more than $20~\mathrm{cm}$ above the table at the 5-second mark. 
Tracking accuracy is computed as the deviation between the measured palm pose and the generated lift pose command in $\mathbb{R}^3$ and $SO(3)$.
As detailed in the third column for each manipulator platform section in \autoref{tab:main_results}, our method achieves an 81.4\% success rate (averaged over 10 random seeds), maintaining an average position error of 4.2 $\mathrm{cm}$ and an orientation error of 15.9$^\circ$ from the generated lift pose command. 
A structured comparison of these results against several baselines is discussed in \autoref{sec:analysis}.

\begin{figure}[t]
    \centering
    \includegraphics[width=0.98\linewidth]{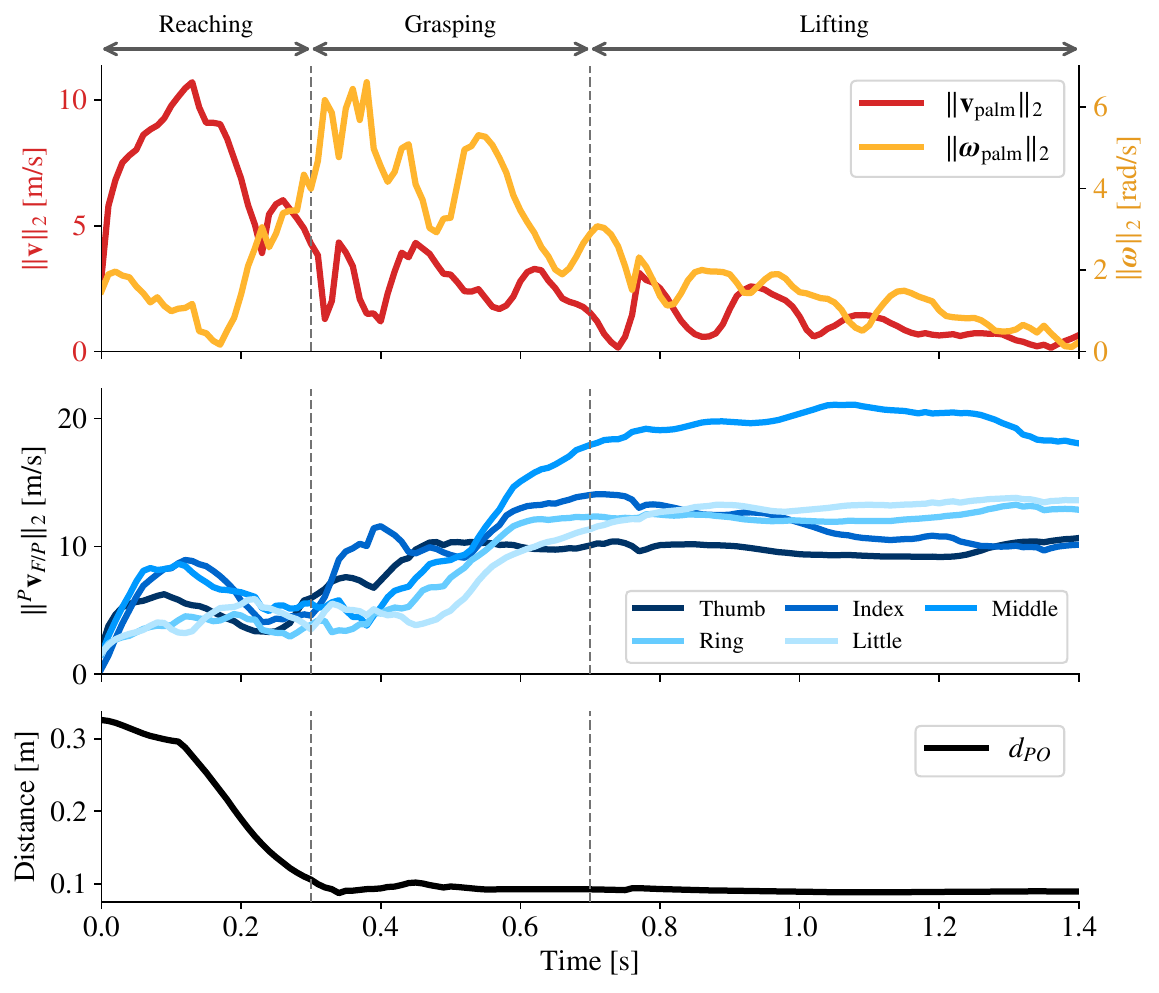}
    \caption{
    Evolution of the arm and hand policy action outputs over time corresponding to the toy airplane execution sequence shown in ~\autoref{fig:sim_results}(a). Phase transitions are intrinsically driven by the decreasing palm-to-object distance (bottom). During reaching, the palm rapidly approaches the object while the fingers remain relatively passive. During grasping, the palm linear velocity drops for precise positioning while the fingertip velocities peak to actively enclose the object. In the lifting stage, the palm translates to the desired pose while the fingers maintain continuous velocity commands to stabilize the grasp.
    }
    \label{fig:action_output}
\end{figure}

To further analyze the dynamic behavior of our control framework, we plot the action outputs during a dexterous grasping task in \autoref{fig:action_output}. 
The figure illustrates the evolution of the desired palm twist and fingertip linear velocities of the 5F hand, which represent the action commands generated by the arm and hand policies, respectively. 
The execution sequence naturally transitions through the three distinct stages of reaching, grasping, and lifting, governed by the decrease in palm-to-object distance.
During the initial reaching stage, the arm executes rapid macroscopic movements to approach the object while the fingertips remain relatively passive. 
As the sequence enters the grasping phase, the palm decelerates to allow for precise local positioning, accompanied by a sharp increase in fingertip velocities as the hand actively encloses the object. 
Finally, during the lifting stage, the palm converges toward the target lift pose while continuous fingertip velocity commands are sustained to ensure a stable and robust grasp.

\begin{figure}[t]
    \centering
    \includegraphics[width=0.98\linewidth]{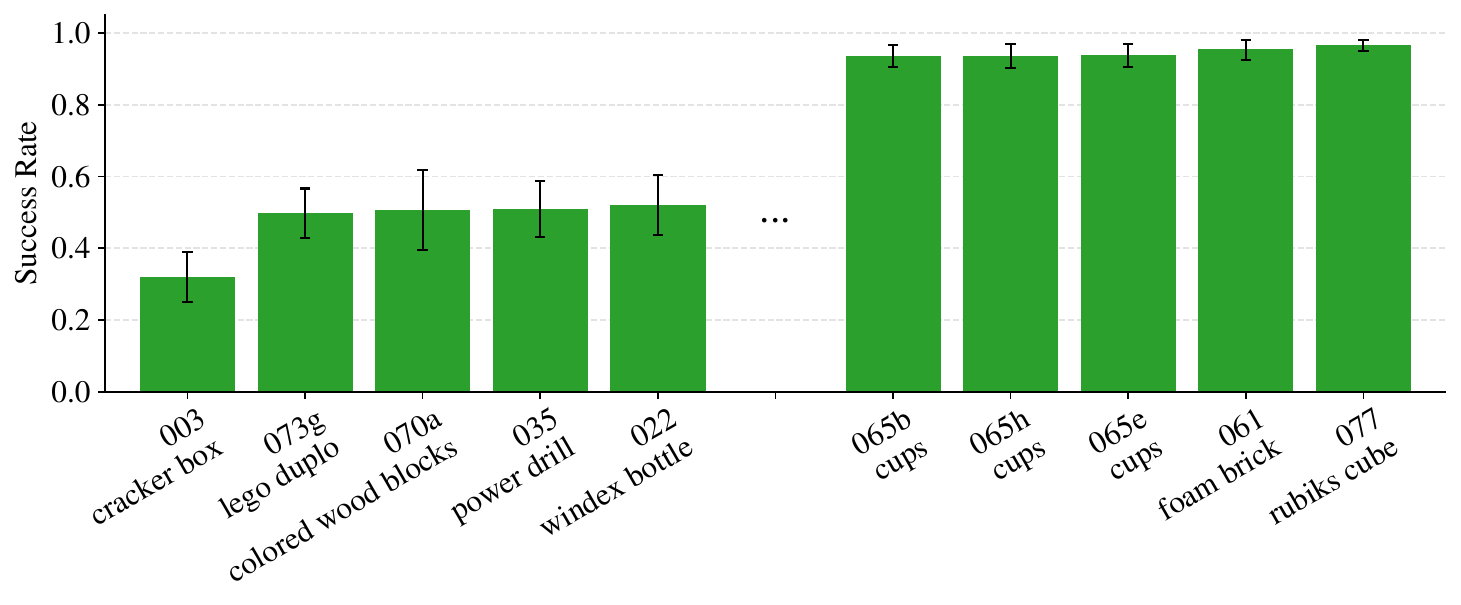}
    \caption{
    Per-object grasp success rates for the 5F hand, highlighting the five lowest-performing (left) and highest-performing (right) objects. 
    The first line of the x-axis labels denotes the corresponding YCB object ID. 
    The results demonstrate that geometrically uniform objects accommodate diverse grasping strategies, yielding higher success rates, whereas objects with extreme aspect ratios or asymmetric geometries demand highly precise and specialized grasps that are harder to execute. 
    }
    \label{fig:per_object_success_rate}
\end{figure}

\autoref{fig:per_object_success_rate} further breaks down the 5F hand's performance by illustrating the top five and bottom five objects in terms of success rate. 
We observe that objects with roughly equal aspect ratios (e.g., cups, foam bricks, Rubik's cubes) achieve the highest success rates, whereas objects with extreme aspect ratios (e.g., cracker boxes, power drills, Windex bottles) remain the most difficult. 
This performance gap suggests that geometrically uniform objects accommodate a wider diversity of successful grasp strategies. 
Conversely, objects with extreme aspect ratios or asymmetric geometries require precise, highly specialized grasping strategies that present a greater challenge for policy learning and execution.

\subsection{Hardware Setup}
\label{subsec:hardware_setup}

\begin{figure}[t]
    \centering
    \includegraphics[width=0.95\linewidth]{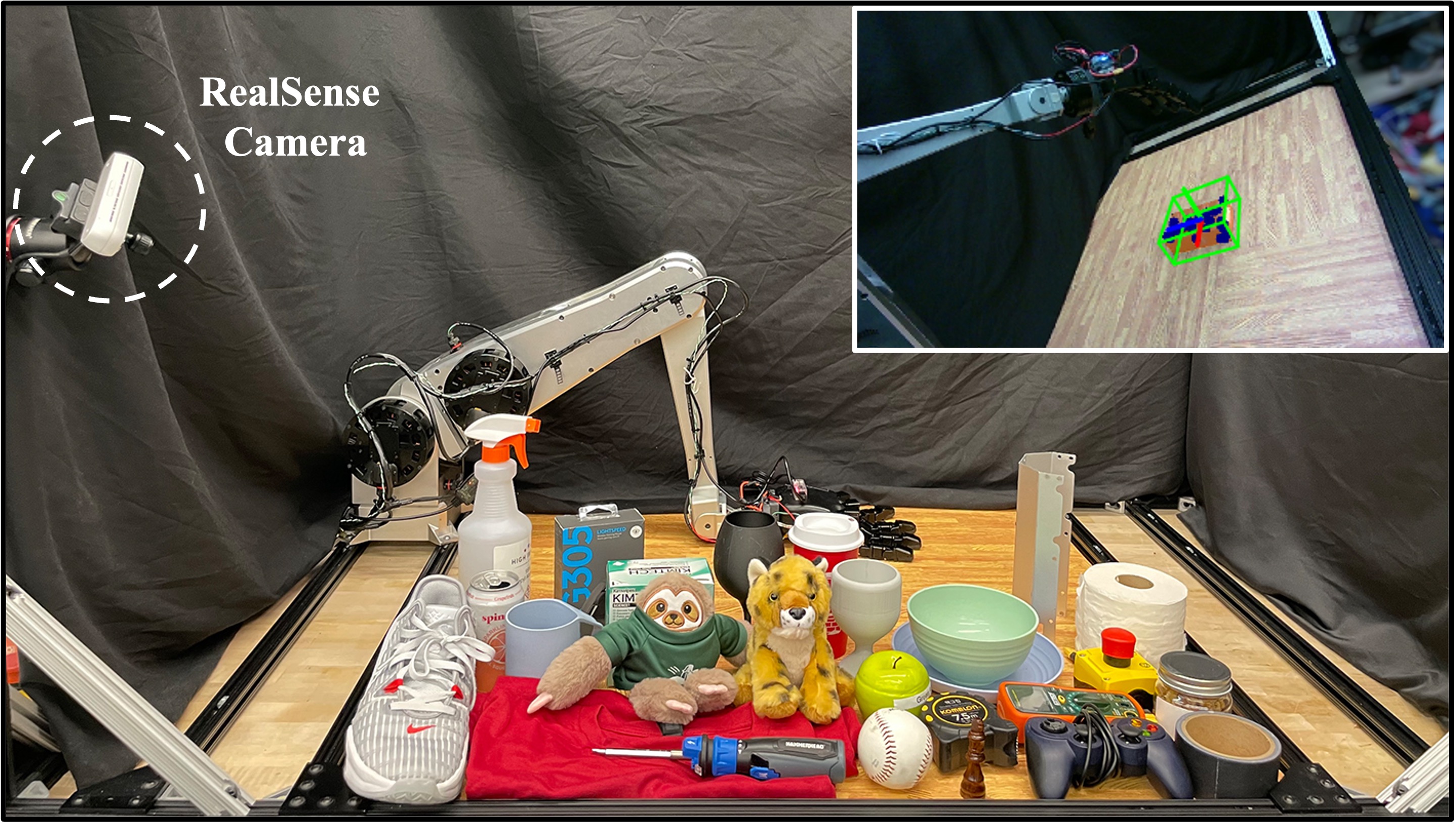}
    \caption{
    Experimental setup for real-world hardware validation. 
    We evaluated our policy on 26 diverse objects. 
    A RealSense camera (top left) computes the object state in real time. The top-right inset shows the camera's perspective, extracting the target object state used as the observation for the RL planner. 
    The manipulator platform is located in the left rear corner.}
    \label{fig:hw_experimental_setup}
\end{figure}

The experimental setup for our real-world hardware validation is illustrated in \autoref{fig:hw_experimental_setup}. 
The workspace consists of a flat tabletop environment measuring approximately $0.8~\mathrm{m} \times 0.9~\mathrm{m}$. 
The manipulator platform, mounted in the left rear corner of the workspace, comprises a 7-DoF robotic arm equipped with a 20-DoF anthropomorphic hand (ROBOTIS HX5-D20).

For visual perception, an Intel RealSense D455 RGB-D camera is mounted at the top left of the workspace, providing a clear view of the table. 
To extract the real-time target object state required as the observation for our RL policy, we utilize the Point2pose algorithm \cite{lin2026point2pose}. 
This 6D pose estimation method allows us to compute the object pose from a single RGB-D image stream. 
As shown in the top-right inset of \autoref{fig:hw_experimental_setup}, the perception system successfully tracks the object state, which is then fed directly to the RL planner. 

\begin{figure*}[tb]
    \centering
    \begin{subfigure}[b]{\textwidth}
        \centering
        \includegraphics[width=\textwidth]{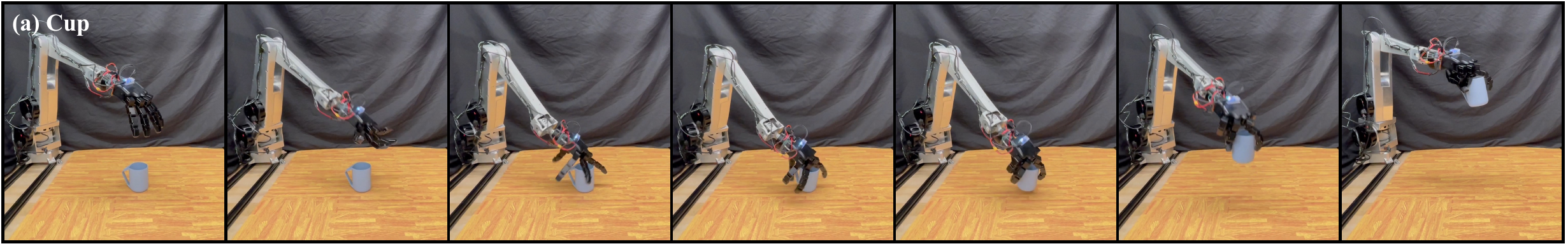}
        \vspace{-3mm} 
    \end{subfigure}
    \begin{subfigure}[b]{\textwidth}
        \centering
        \includegraphics[width=\textwidth]{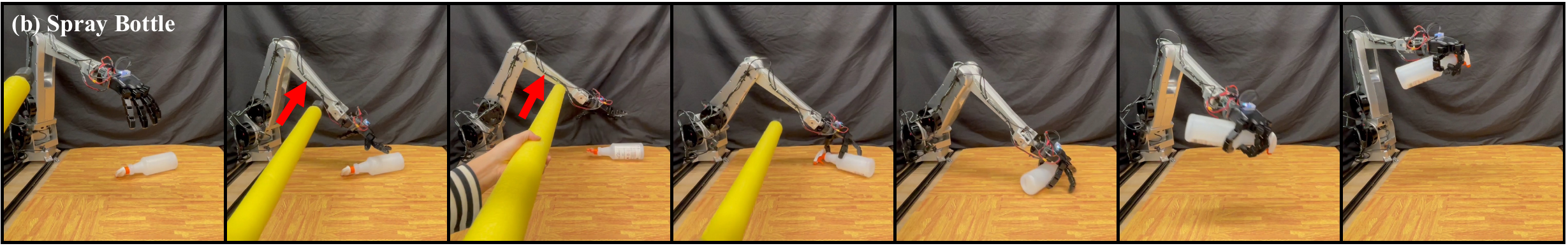}
        \vspace{-3mm} 
    \end{subfigure}
    \begin{subfigure}[b]{\textwidth}
        \centering
        \includegraphics[width=\textwidth]{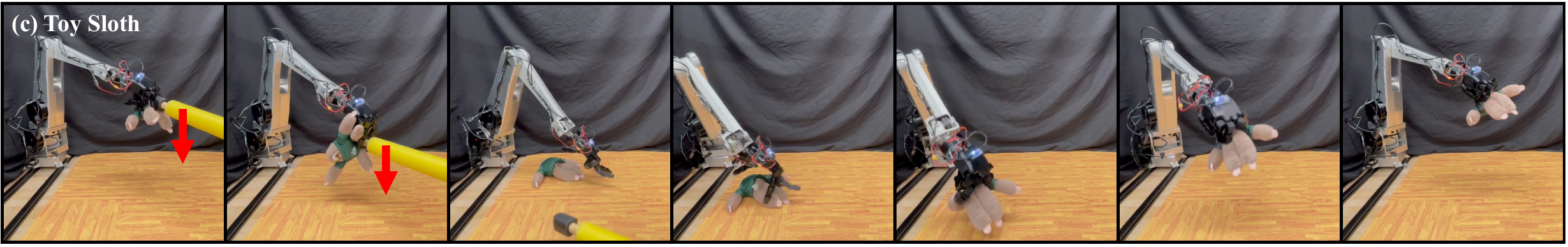}
        \vspace{-3mm} 
    \end{subfigure}
    \caption{
    Hardware experiments demonstrating real-world grasping with the 5F hand on previously unseen objects. 
    (a) Grasping a cup demonstrates the framework's generalization capability to novel object geometries. 
    (b) Grasping a spray bottle illustrates the reactive control robustness of the proposed framework: after an unexpected external disturbance pushes the robot arm away during the reaching phase, the manipulator recovers its target pose and successfully executes the grasp. 
    (c) Re-grasping a deformable toy sloth showcases the system's dynamic adaptability. When an external disturbance physically displaces the object from the initial grasp, the robot arm continuously recalculates its task-space trajectory to successfully recover and re-execute the grasp.
    }
    \label{fig:hw_results}
\end{figure*}

\subsection{Hardware Results}
\label{subsec:hardware_results}

To validate the real-world applicability and zero-shot generalization of our proposed framework, we evaluated the 5F hand setup on a diverse set of 26 everyday items (see~\autoref{fig:hw_experimental_setup}). 
All 26 objects were completely unseen during the training phase, and the test set included several challenging deformable objects. 
Our method successfully grasped and lifted 22 out of the 26 objects within a few attempts. 
These results demonstrate that the policies trained purely in simulation can robustly transfer to physical hardware without the need for real-world fine-tuning. 

The qualitative performance of the framework is further detailed in \autoref{fig:hw_results}, which highlights three representative execution sequences. 
\autoref{fig:hw_results}(a) illustrates the successful grasping of a cup, demonstrating the framework's ability to smoothly adapt finger coordination to unseen geometries. 
More importantly, \autoref{fig:hw_results}(b) showcases the robust reactive control capabilities of the proposed system. 
While the manipulator was approaching a spray bottle, an unexpected external physical disturbance was applied to the robot arm. 
Because the policy continuously operates as a high-frequency reactive closed-loop system, the arm and hand quickly recovered from the disturbance, adjusted their approach trajectory on the fly, and successfully executed the grasp.
Similarly, \autoref{fig:hw_results}(c) demonstrates a successful re-grasp of a deformable toy sloth after a physical disturbance displaced the object and broke the initial grasp; the system continuously recalculated its task-space trajectory to recover the object and complete the lift.

Despite the high overall success rate, we observed consistent failure cases on four specific items: a flat plate, a thin screwdriver, a small chess piece, and a slippery metal sheet. 
These failures provide valuable insight into the physical limitations of the current setup. 
The slippery plate was difficult to lift due to a lack of sufficient friction and the absence of tactile feedback in our current observation space. 
Meanwhile, the extreme geometric properties of the flat plate, thin screwdriver, and small chess piece made them difficult to enclose securely.

Full motion sequences for both simulation and hardware experiments are available in the supplementary video.

%% file: Sections/08_analysis.tex
\section{Analysis}
\label{sec:analysis}

In this section, we present an in-depth analysis of our proposed hybrid hierarchical control framework, highlighting the unique advantages afforded by our task-space action representation and the integration of a low-level QP controller. 
The remainder of this section is structured into four parts. 
First, we evaluate the architectural benefits of our hierarchical, multi-agent RL approach. 
Second, we examine the critical role of the QP controller in enforcing kinematic limitations and ensuring safe execution. 
Third, we demonstrate the adaptability and steerability of our framework across different operational requirements. 
Finally, we conclude with a holistic discussion on the implications of these design choices for achieving robust and reactive dexterous grasping.

\begin{figure*}[tb]
    \centering
    \includegraphics[width=0.8\textwidth]{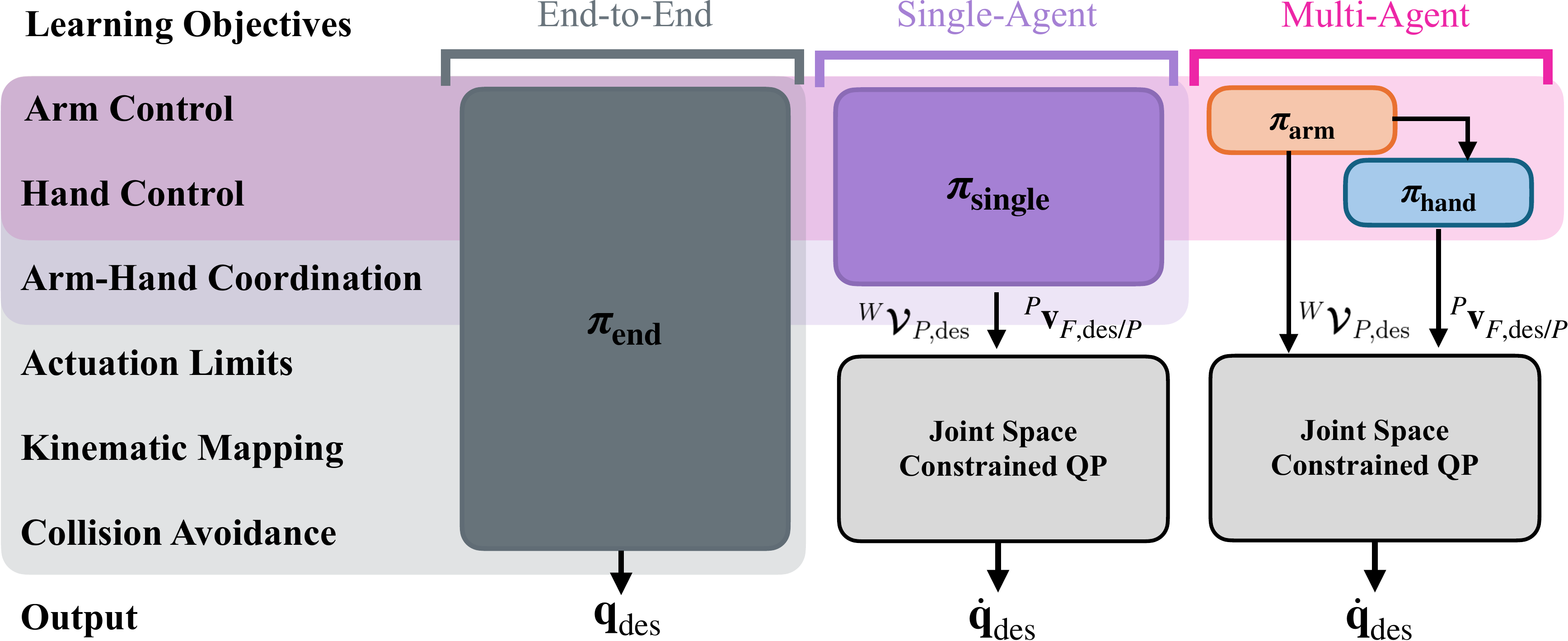}
    \caption{
    Comparison of control architectures and corresponding learning objectives. 
    Unlike the \etoe~approach, which must implicitly learn the robot's differential kinematics (Kinematic Mapping) and low-level physical constraints, our hierarchical decomposition delegates these complexities to a joint-space constrained QP controller. 
    Furthermore, by separating the policy into specialized arm and hand agents, we remove the burden of learning complex arm-hand coordination from a single planner. 
    This architectural shift allows our policies to focus exclusively on discovering high-level manipulation strategies, leading to more efficient and targeted learning. 
    }
    \label{fig:architectural_comparison}
    \vspace{-2mm}
\end{figure*}

\subsection{Architectural Analysis of the RL Framework}
\label{subsec:rl_architecture}

In this section, we evaluate the architectural benefits of our hierarchical, \multiagent~RL approach by comparing it against two baseline formulations: an \etoe~architecture and a hierarchical \singleagent~architecture (\autoref{fig:architectural_comparison}). 
Dexterous grasping inherently demands the simultaneous management of multiple complex objectives: controlling arm and hand joints, coordinating their relative motions, learning the kinematic mapping from joint space to task space, respecting physical hardware constraints, and avoiding environmental collisions. 
Requiring an \etoe~policy to learn all these aspects simultaneously from scratch is notoriously sample-inefficient and prone to suboptimal convergence. 
Our hierarchical formulation mitigates this by delegating explicit differential kinematics and physical constraint enforcement to a low-level QP controller. 
Furthermore, by partitioning the high-level RL planner into specialized arm and hand agents, we decouple the complex arm-hand coordination problem. 
This decomposition effectively mitigates the curse of dimensionality, allowing each policy to focus exclusively on discovering its respective high-level manipulation strategies.

\begin{figure}[t]
    \centering
    \includegraphics[width=0.98\linewidth]{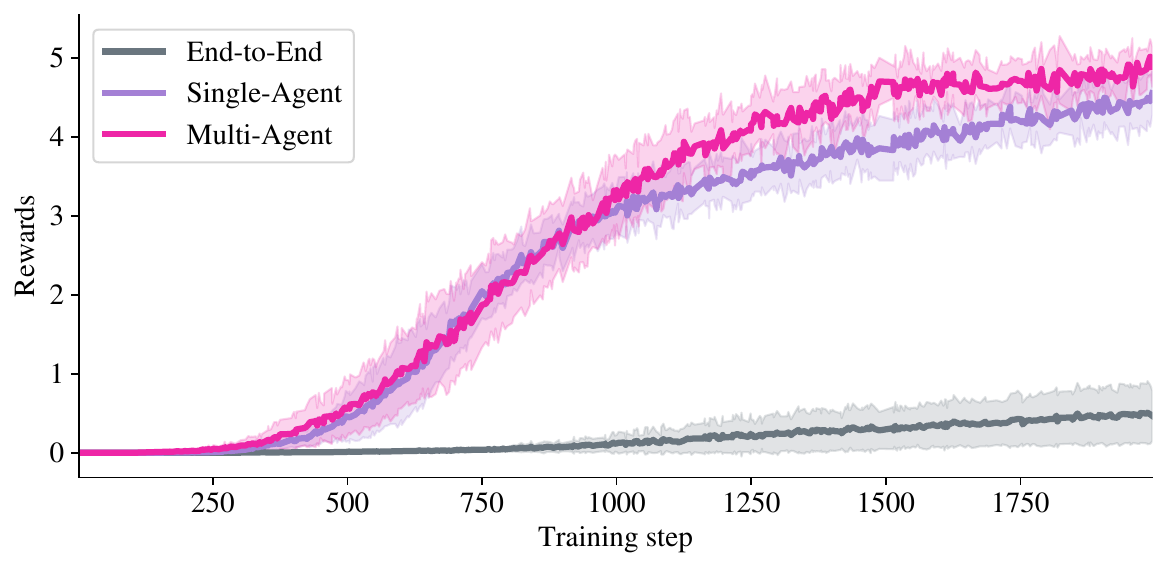}
    \caption{
    Training performance comparison of the proposed framework against two baseline architectures. 
    The plot displays the goal-conditioned task-space lift reward averaged across 10 random seeds. 
    Our \multiagent~approach converges to the highest final reward, whereas the \etoe~baseline struggles to learn the complex grasping task entirely.
    }
    \label{fig:reward_curve}
\end{figure}

We first evaluate the learning efficiency and policy performance of these architectures by comparing their task reward evolution during training, averaged across 10 random seeds (\autoref{fig:reward_curve}). 
The learning curves clearly indicate that delegating low-level control to a QP controller (as seen in both the \singleagent~and \multiagent~frameworks) yields significantly faster convergence and higher asymptotic rewards compared to the \etoe~baseline. 
Notably, for the highly dexterous 5F hand, further decomposing the RL planner into two specialized agents allows the \multiagent~architecture to achieve an even higher final reward than the \singleagent~approach. 
This demonstrates that lowering the learning burden on the neural network and reducing the action space, by explicitly separating the roles of arm and hand, substantially improves learning efficiency and performance for high-DoF systems.

As shown in \autoref{tab:main_results}, the \multiagent~architecture achieves the highest success rate (81.4\%) and the lowest pose errors for the 20-DoF 5F hand. 
Interestingly, while the \singleagent~baseline achieves a slightly shorter time to success for the 5F hand, it does so by drastically sacrificing pose command accuracy, as evidenced by its high orientation error (59.7$^\circ$). 
Conversely, for the simpler 8-DoF 2F gripper, the \singleagent~architecture remains highly effective due to the lower-dimensional action space of the fingertips, suggesting that the optimal RL architecture is intrinsically tied to the morphological complexity of the end-effector; however, the \multiagent~approach consistently provides superior tracking performance in terms of both position and orientation error across all platforms.

\begin{table*}[t] 
    \centering
    \caption{Quantitative Performance Comparison Across Policy Architectures and Manipulator Platforms}
    \label{tab:main_results}
    \begin{tabular}{@{}l ccc c ccc@{}}
        \toprule
        \multirow{2}{*}[-0.7ex]{\textbf{Metric}}
        & \multicolumn{3}{c}{\textbf{5F Hand}} & & 
        \multicolumn{3}{c}{\textbf{2F Gripper}} \\
        \cmidrule(lr){2-4} \cmidrule(l){6-8} 
        
        & End-to-End & Single-Agent & Multi-Agent & & End-to-End & Single-Agent & Multi-Agent \\
        \midrule
        
        Success Rate [$\%$] $\uparrow$ & $13.2 \pm 12.8$ & $80.9 \pm 4.6$ & $\mathbf{81.4 \pm 3.3}$ && $60.3 \pm 8.7$ & $\mathbf{81.7 \pm 2.9}$ & $77.2 \pm 2.4$ \\
        Time to Success [$\mathrm{s}$] $\downarrow$ & $1.257 \pm 0.891$ & $\mathbf{1.078 \pm 0.050}$ & $1.196 \pm 0.053$ && $\mathbf{0.913 \pm 0.167}$ & $1.009 \pm 0.074$ & $1.167 \pm 0.060$ \\
        Position Error [$\mathrm{cm}$] $\downarrow$ & $21.5 \pm 10.5$ & $5.8 \pm 0.5$ & $\mathbf{4.2 \pm 0.3}$ && $6.3 \pm 0.8$ & $3.9 \pm 0.3$ & $\mathbf{3.8 \pm 0.3}$ \\
        Orientation Error [$\mathrm{deg}$] $\downarrow$ & $104.1 \pm 32.0$ & $59.7 \pm 38.5$ & $\mathbf{15.9 \pm 1.0}$ && $37.6 \pm 14.2$ & $12.9 \pm 1.0$ & $\mathbf{9.9 \pm 0.7}$ \\
        
        \bottomrule
    \end{tabular}
\end{table*}

\subsection{Safety and Constraint Enforcement via QP Control}
\label{subsec:qp_controller}

In this section, we evaluate the efficacy of the low-level QP controller in enforcing physical constraints while executing the task-space action generated by the high-level RL policy, which serves as the desired velocity inputs to the controller. 
We analyze this constraint enforcement from two distinct perspectives: first, the effect of the action magnitude on tracking error and constraint violations; second, the spatial distribution of the task-space velocity feasibility envelope.

\begin{figure}[t]
    \centering
    \includegraphics[width=0.9\linewidth]{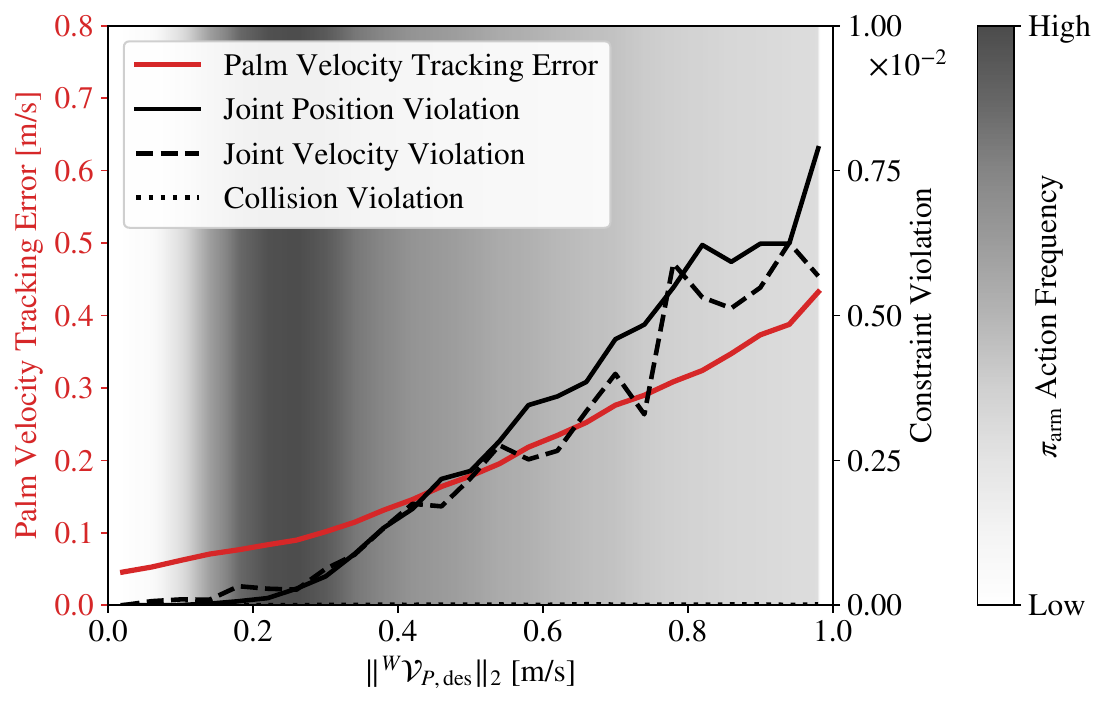}
    \caption{
    Task-space velocity tracking performance and constraint violation for the arm policy. 
    As the desired palm velocity ($||{}^W\twist_{P,\text{des}}||_2$) increases, the mean tracking error (solid red line) grows, predominantly due to joint limit enforcement by the QP controller. 
    However, the background density reveals that the RL policy's actions congregate almost entirely below $0.3~\mathrm{m/s}$, where the framework accurately tracks the desired velocities while strictly satisfying all physical constraints.
    Joint position (solid black line), joint velocity (dashed black line), and collision (dotted black line) violations correspond to the constraints defined in \autoref{eq:qp_pos_limit}, \autoref{eq:qp_vel_limit}, and \autoref{eq:qp_collision}, respectively.
    }
    \label{fig:qp_error}
\end{figure}

We first examine the magnitude of the arm policy's actions across multiple grasping tasks in \autoref{fig:qp_error}. 
As the norm of the desired palm twist $||{}^W\twist_{P,\text{des}}||_2$ increases, the mean tracking error (solid red line) grows correspondingly. 
To identify the source of this tracking deviation, we analyze the constraint violation defined in our QP formulation: joint position limits (solid black line) \eqref{eq:qp_pos_limit}, joint velocity limits (dashed black line) \eqref{eq:qp_vel_limit}, and collision avoidance (dotted black line) \eqref{eq:qp_collision}.
The figure reveals that the deviation is predominantly driven by the activation of joint limits constraints, which safely clamp infeasible desired velocities to prevent hardware damage. 
By overlaying the action frequency distribution of the RL policy (grayscale background), we observe that the policy naturally learns to operate within a highly executable regime below 0.3~$\mathrm{m/s}$. 
In this heavily populated region, the QP controller achieves highly accurate tracking with no constraint violations, demonstrating a successful synergy between the RL agent's learned behavior and the strict physical limits enforced by the lower layer controller.

\begin{figure}[t]
    \centering
    \includegraphics[width=\linewidth]{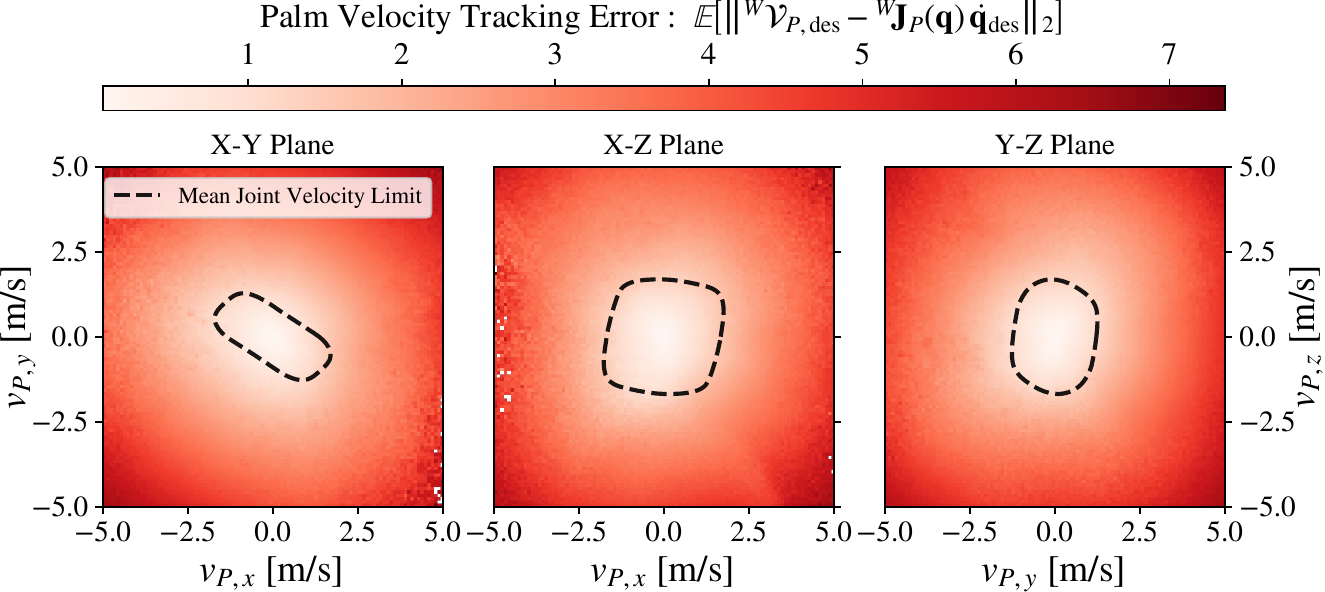}
    \caption{
    Spatial distribution of the palm velocity tracking error across the task-space Cartesian planes. 
    The colormap illustrates the mean tracking error magnitude, while the dashed line outlines the mean joint velocity limit mapped into the operational space. 
    The region of high tracking accuracy (white) spatially conforms to the geometric shape of the joint velocity limit envelope. 
    This spatial alignment confirms that the tracking performance is strictly governed by the physical capabilities of the manipulator, demonstrating that the error growth observed in~\autoref{fig:qp_error} is a direct consequence of the QP controller enforcing these complex, non-linear kinematic boundaries.
    }
    \label{fig:qp_error_spatial}
\end{figure}

While \autoref{fig:qp_error} establishes that tracking error scales with the magnitude of the action, it does not capture the directional limitations dictated by the robot's kinematics. 
To investigate this spatial relationship, we profile the palm velocity tracking error across the Cartesian task-space planes in \autoref{fig:qp_error_spatial}. 
By mapping the mean joint velocity limits into the operational space (dashed boundary, see Appendix~\ref{appendix:joint_vel_contours} for derivation) and overlaying them onto the tracking error heatmap, a clear geometric correlation emerges between kinematic limits and tracking performance. 
The high-accuracy tracking region, represented by the central white area, distinctly mirrors the complex, non-linear shape of the physical limit envelope in each respective plane.

This spatial alignment demonstrates that the tracking performance is strictly and dynamically governed by the manipulator's true kinematic boundaries. 
Together, these results validate the safety constraints enforced by our hierarchical architecture. 
The RL policy is permitted to generate task-space actions without requiring an explicit model of the robot's differential kinematics. 
Simultaneously, the QP controller acts as a rigorous, deterministic safety filter, ensuring that the executed motions conform to the hardware's operational space limits.

In contrast, we observed that the \etoe~baseline, which relies on reward-based constraint enforcement, suffers from frequent early terminations during the exploration phase due to joint limit saturation and environment collisions. 
This prevents the policy from collecting the high-quality transition data necessary for successful convergence.
By decoupling safety enforcement from the learning signal, our hierarchical approach allows the RL agent to explore the task space more effectively while the QP controller ensures the safety and feasibility of executed motions.

\begin{figure}[t]
    \centering
    \includegraphics[width=0.95\linewidth]{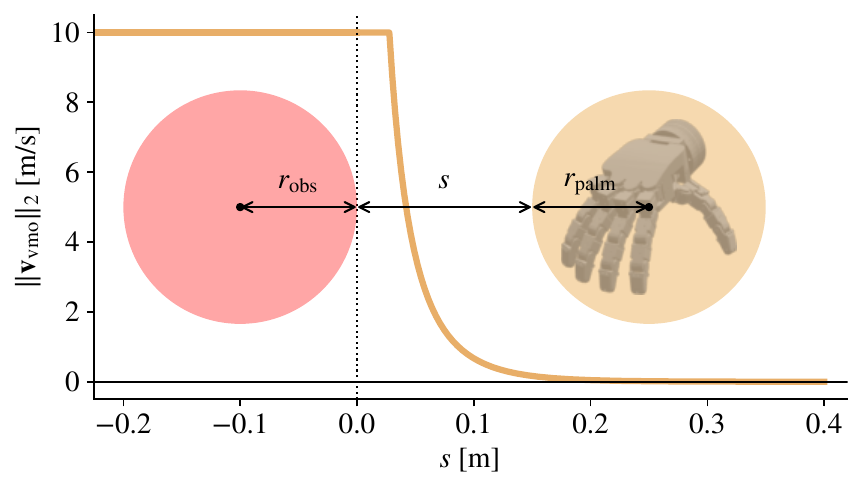}
    \caption{
    Design of the APF for online task-space velocity modulation. 
    To evaluate the framework's task-space steerability, we formulate a custom repulsive velocity field. 
    When the center of the palm approaches within $10~\mathrm{cm}$ of an obstacle's surface ($s \leq 0.1~\mathrm{m}$), the APF generates a radial outward velocity $\mathbf{v}_{\text{vmo}}$ to be added to the arm agent's action. 
    The magnitude scales inversely with the separation distance and is strictly capped at $10~\mathrm{m/s}$ to ensure stability when passed to the low-level QP controller.    
    }
    \label{fig:vmo_eqn}
\end{figure}

\subsection{Framework Adaptability and Steerability}
\label{sec:framework_adaptability}

\begin{figure*}[tb]
    \centering
    \includegraphics[width=0.95\textwidth, trim={0 0 0 4mm}, clip]{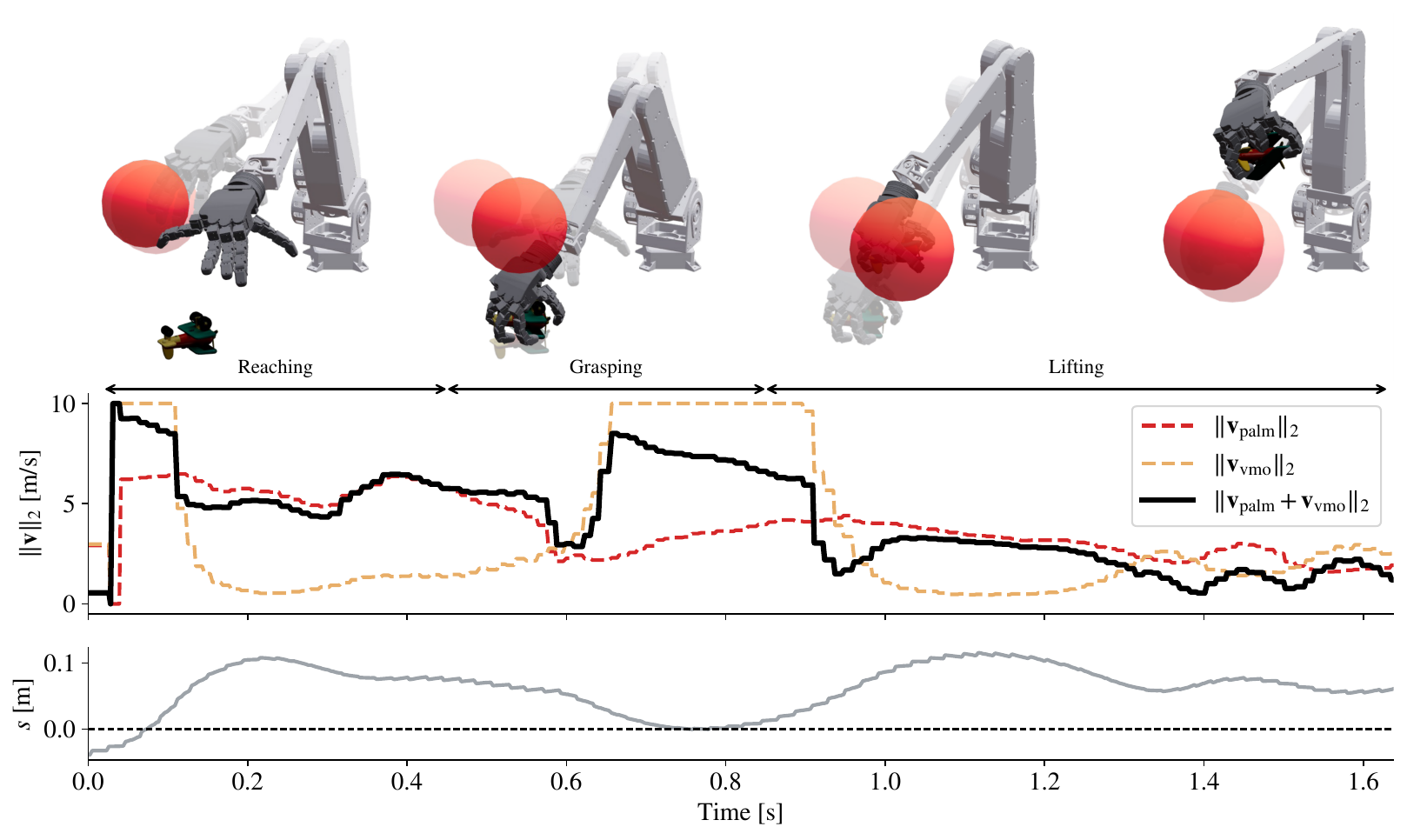}
    \caption{
    By superimposing the APF-generated repulsive velocity ($\mathbf{v}_{\text{vmo}}$) onto the RL planner's output ($\mathbf{v}_{\text{palm}}$), the manipulator dynamically avoids an unseen obstacle (red sphere) zero-shot. 
    \textbf{(Top)} Sequence of collision-free motion across the reaching, grasping, and lifting phases. \textbf{(Middle)} Evolution of the velocity norms, showing the resultant task-space command (black solid line) smoothly adapting to the obstacle's presence. \textbf{(Bottom)} The separation distance $s$ is safely maintained at or above zero, validating successful collision avoidance without interrupting the grasp.
    }
    \label{fig:vmo}
\end{figure*}

\begin{figure}[t]
    \centering
    \includegraphics[width=0.98\linewidth]{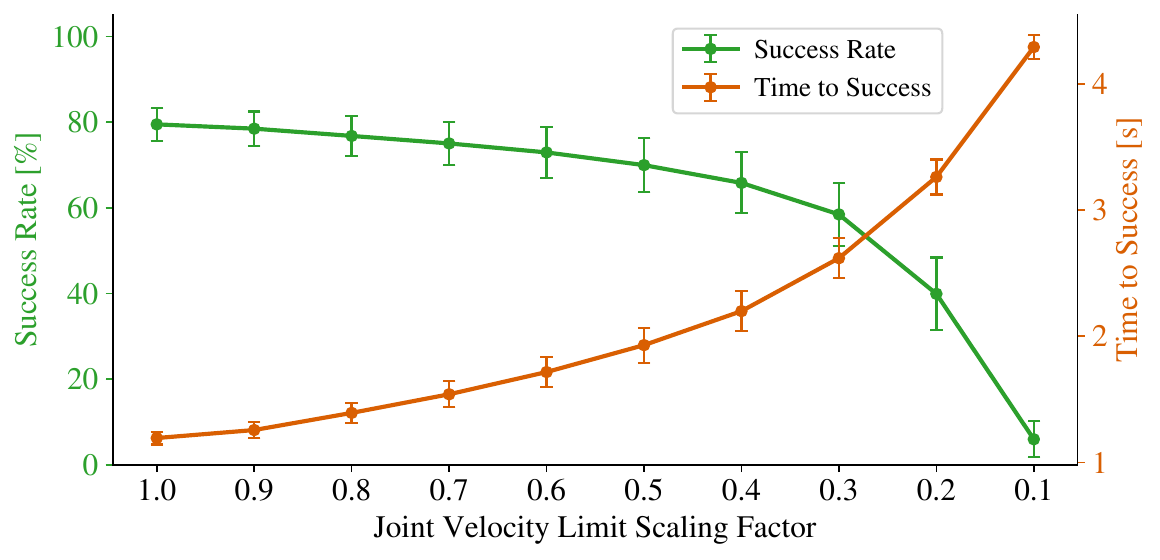}
    \caption{
    Post-training steerability via joint velocity limit scaling. 
    Reducing the arm's default joint velocity limits (2 rad/s) on the low-level QP controller smoothly increases the time-to-success while maintaining robust success rates down to a 0.5 scaling factor. 
    This demonstrates the framework's ability to dynamically adjust operational speed and safety trade-offs without requiring policy retraining.
    }
    \label{fig:vel_factor}
\end{figure}

A fundamental limitation of monolithic RL policies is their rigidity post-training; adapting to unseen obstacles or new safety limits typically requires expensive retraining because the network implicitly memorizes the system's exact kinematic response. 
By explicitly decoupling task-space intent from joint-space execution, our hierarchical architecture overcomes this bottleneck. 
This modularity enables zero-shot adaptability and steerability through two complementary mechanisms: online velocity modulation in the task space, and dynamic constraint adjustments in the joint space.

First, because the high-level policy outputs task-space velocity commands, its intent can be intuitively modulated online using classical spatial techniques. 
To demonstrate this, we designed a custom APF for reactive obstacle avoidance (\autoref{fig:vmo_eqn}). 
When the palm approaches within $10~\mathrm{cm}$ of an obstacle, the APF generates a repulsive radial velocity ($\mathbf{v}_{\text{vmo}}$) that is directly superimposed onto the arm agent's action. 
As detailed in \autoref{fig:vmo}, this velocity modulation allows the manipulator to smoothly deviate from its original trajectory to skirt a dynamic obstacle ($s \geq 0$). 
Crucially, the resultant task-space command safely guides the robot around the intrusion without interrupting the high-level grasping logic or requiring the RL agent to explicitly learn obstacle avoidance.

Second, our framework allows the degree of execution conservativeness to be adjusted post-training via the low-level QP controller.  
While altering actuator limits post-training generally breaks monolithic policies due to a sudden mismatch in transition dynamics, our architecture naturally handles these changes. 
As shown in \autoref{fig:vel_factor}, systematically reducing the default arm joint velocity limits within the QP constraints smoothly increases the time-to-success while maintaining robust grasping performance down to a 0.5 scaling factor. 
This demonstrates that system operators can dynamically dial along a speed-safety trade-off curve on the fly, relying entirely on the QP to synthesize strictly feasible joint velocity commands for the newly constrained envelope.

Together, this zero-shot adaptability and steerability post-training, both in task space and joint space, highlight the unique advantages of interfacing learning-based planners with physics-based controllers.

\subsection{Discussion}
\label{sec:discussion}

To contextualize the empirical and architectural findings of this work, we address several fundamental questions arising from the design and applicability of our hybrid hierarchical control framework for reactive dexterous grasping.

\noindent \textbf{When is a hierarchical, multi-agent RL architecture necessary?} \\
Our empirical results reveal that the optimal RL architecture is intrinsically tied to the morphological complexity of the end-effector. 
For lower-complexity platforms, such as the 8-DoF 2F gripper, a single unified RL agent is sufficient and often more efficient, as the coordination burden is minimal. 
However, as demonstrated by our 20-DoF 5F hand experiments, scaling to highly dexterous platforms introduces a severe curse of dimensionality. 
Forcing a single neural network to untangle the highly coupled coordination of arm reaching and multi-finger grasping severely hinders learning. 
In these high-DoF scenarios, explicitly separating the roles of the arm and hand into a multi-agent architecture offers clear performance advantages. 
This decoupling isolates the action representations, resolves coordination complexity, and enables the policies to master intricate manipulation strategies that a single agent struggles to discover.

\noindent \textbf{Why delegate constraint enforcement to a low-level QP instead of the RL reward function?} \\
In standard end-to-end RL paradigms, safety limits and kinematic constraints are typically handled via soft penalty terms in the reward function. 
However, balancing these safety penalties with primary task rewards often creates competing objectives, which reduces overall training efficiency.
By removing these constraints from the RL objective and delegating their enforcement to a low-level QP controller, we significantly reduce the learning burden on the policy, allowing it to focus exclusively on discovering high-level manipulation strategies.
Moreover, while soft penalties encourages safe behavior during training, they cannot deterministically guarantee hardware safety during real-world execution, especially when the policy encounters out-of-distribution observations. 
By delegating constraint enforcement to a low-level QP controller, our framework establishes strict physical boundaries (as visualized in \autoref{fig:qp_error_spatial}). 
Finally, delegating these constraints to the QP controller unlocks zero-shot steerability.
System operators can dynamically adjust the QP velocity limits at runtime to directly control operational speed and dictate safety margins, allowing the RL policy to continue functioning robustly without requiring any retraining.

\noindent \textbf{What makes the proposed framework highly reactive to real-world disturbances?} \\
The robustness observed in our hardware experiments, particularly the reactive recovery during the spray bottle grasping task, stems from successfully combining the strengths of RL-based spatial reasoning with high-frequency model-based control. 
Because the RL policies output task-space desired velocities rather than low-level joint torques, they operate purely on the relative spatial relationship between the manipulator and the object. 
When an unexpected external disturbance physically displaces the robot arm, the RL policy simply observes the new relative pose and continuously generates corrective task-space actions to close the gap. 
Concurrently, the high-frequency QP controller ensures that these rapid corrective commands are instantly translated into safe, kinematically feasible joint motions. 
This structural isolation, leveraging RL for high-level adaptive robustness and the QP for low-level feasibility, is what ultimately enables the reactive disturbance recovery in unpredictable physical environments.

%% file: Sections/09_conclusion.tex
\section{Conclusion}

In this work, we presented a hybrid hierarchical control framework for reactive dexterous grasping that explicitly decouples high-level spatial intent from low-level joint execution. 
At the high level, a multi-agent RL architecture, comprising specialized arm and hand agents, operates as a spatial planner to generate desired task-space velocity commands. 
These commands are subsequently converted to feasible joint velocity commands by a GPU-parallelized QP controller. 
By delegating differential kinematics, collision avoidance, and physical constraint enforcement to the model-based controller, the proposed framework isolates the RL policy to task-level reasoning. 
Consequently, this structural decoupling effectively achieves robust and reactive dexterous grasping performance and accelerates training convergence, while strictly ensuring hardware safety.


Extensive validation across large-scale simulations and real-world hardware deployments demonstrated the robustness and adaptability of the proposed framework. 
Executed on a 7-DoF arm equipped with a 20-DoF anthropomorphic hand, the policies achieved successful zero-shot transfer to a diverse set of previously unseen objects. 
Our analysis confirmed that the structural isolation of task-space action representation from joint-space execution enables fluid, reactive recovery from unexpected external physical disturbances. 
Additionally, this decoupling unlocks zero-shot steerability, allowing system operators to dynamically adjust safety boundaries and smoothly avoid dynamic obstacles without requiring any policy retraining.

While the current framework has proven highly effective for robust reactive grasping, future work will focus on extending this architecture to a broader spectrum of complex manipulation tasks. 
Specifically, we aim to transition from static grasping to dynamic, continuous contact scenarios, applying our hierarchical multi-agent formulation to tasks such as dexterous in-hand manipulation and multi-stage pick-and-place operations. 
By building upon this robust integration of RL and model-based control, we intend to further bridge the gap toward generalized dexterous manipulation in highly unstructured environments.

%% file: Sections/10_appendix.tex
{
\appendices



\section{Multi-Agent RL Formulation Details}

This section provides the comprehensive implementation details and parameter configurations for the proposed hierarchical RL framework. 
\autoref{tab:arm_obs} and \autoref{tab:hand_obs} detail the observation space dimensions for the arm and hand agents, respectively.
\autoref{tab:reward_weights} lists the reward weights for both policies, while \autoref{tab:domain_randomization} defines the ranges used for domain randomization. 
Finally, the PPO hyperparameters employed during the training phase are summarized in \autoref{tab:ppo_hyperparameters}.

\begin{table}[h]
\centering
\caption{Arm Agent Observation Space Dimensions}
\label{tab:arm_obs}
\renewcommand{\arraystretch}{1.3} 
\begin{tabular}{l c c c}
\toprule
\textbf{Observation Component} & \textbf{Symbol} & \textbf{2F Gripper} & \textbf{5F Hand} \\
\midrule
Joint Positions       & $\mathbf{q}$                        & 15 & 27 \\
Joint Velocities      & $\dot{\mathbf{q}}$                  & 15 & 27 \\
Object Position       & $\mathbf{p}_{\text{obj}}$           & 3  & 3  \\
Object Orientation    & $\mathbf{Q}_{\text{obj}}$           & 4  & 4  \\
Object Dimensions     & $\mathbf{d}_{\text{obj}}$           & 3  & 3  \\
Lift Pose Command     & $\mathbf{x}_{\text{cmd}}$           & 7  & 7  \\
Previous Arm Action   & $\mathbf{a}_{\text{arm}}^{t-1}$     & 6  & 6  \\
\midrule
\textbf{Total}        & $\mathcal{O}_{\text{arm}}$          & \textbf{53} & \textbf{77} \\
\bottomrule
\end{tabular}
\end{table}

\begin{table}[h]
\centering
\caption{Hand Agent Observation Space Dimensions}
\label{tab:hand_obs}
\renewcommand{\arraystretch}{1.3} 
\begin{tabular}{l c c c}
\toprule
\textbf{Observation Component} & \textbf{Symbol} & \textbf{2F Gripper} & \textbf{5F Hand} \\
\midrule
Hand Joint Positions       & $\mathbf{q}_{\text{hand}}$                 & 8  & 20 \\
Hand Joint Velocities      & $\dot{\mathbf{q}}_{\text{hand}}$           & 8  & 20 \\
Previous Hand Action       & $\mathbf{a}_{\text{hand}}^{t-1}$           & 6  & 15 \\
Object Position            & ${}^{P}\mathbf{p}_{\text{obj}}$            & 3  & 3  \\
Object Orientation         & ${}^{P}\mathbf{Q}_{\text{obj}}$            & 4  & 4  \\
Object Dimensions          & $\mathbf{d}_{\text{obj}}$                  & 3  & 3  \\
Palm Spatial Twist         & ${}^{P}\boldsymbol{\mathcal{V}}_{P}$       & 6  & 6  \\
Current Arm Action         & $\mathbf{a}_{\text{arm}}^{t}$              & 6  & 6  \\
\midrule
\textbf{Total}             & $\mathcal{O}_{\text{hand}}$                & \textbf{44} & \textbf{77} \\
\bottomrule
\end{tabular}
\end{table}

\begin{table}[h]
    \centering
    \caption{Reward Weights in Arm and Hand Policies}
    \label{tab:reward_weights}
    \begin{tabular}{@{}llc@{}}
        \toprule
        \textbf{Policy} & \textbf{Reward Term} & \textbf{Weight} \\
        \midrule
        \multirow{7}{*}{\textbf{Arm}}
        & Palm-to-Object Euclidean Distance & 1.0 \\ 
        & Palm Orientation Alignment to Palm-to-Object Vector & 1.0 \\ 
        & Stable Grasp & 1.0 \\ 
        & Goal-Conditioned Task-Space Lift & 10.0 \\ 
        \cmidrule{2-3}
        & Action Smoothness 1st-order & 2e-5 \\ 
        & Action Smoothness 2nd-order & 2e-6 \\ 
        & Termination & 100 \\ 
        \midrule[\heavyrulewidth] 
        \multirow{4}{*}{\textbf{Hand}}
        & Stable Grasp & 2.0 \\ 
        \cmidrule{2-3}
        & Action Smoothness 1st-order & 1e-5 \\ 
        & Action Smoothness 2nd-order & 1e-6 \\ 
        & Termination & 100 \\ 
        \bottomrule
    \end{tabular}
\end{table}

\begin{table}[h]
    \centering
    \caption{Domain Randomization Parameters}
    \label{tab:domain_randomization}
    \begin{tabular}{@{}llc@{}}
        \toprule
        \multicolumn{1}{c}{\textbf{Component}} & \textbf{Parameter} & \textbf{Range} \\
        \midrule
        
        \multirow{2}{*}{\textbf{Robot Properties}}
        & Link Mass & $\mathcal{U}(0.95, 1.05) \times \text{Default}$ \\ 
        & Surface Friction ($\mu_s, \mu_d$) & $\mathcal{U}(0.6, 1.0)$ \\ 
        \midrule
        
        \multirow{2}{*}{\textbf{Object Properties}} 
        & Mass & $\mathcal{U}(0.8, 1.2) \times \text{Default}$ \\ 
        & 3D Scale ($x, y, z$) & $\mathcal{U}(0.8, 1.2) \times \text{Default}$ \\ 
        
        \bottomrule
    \end{tabular}
\end{table}

\begin{table}[h]
    \centering
    \caption{PPO Hyperparameters for Modular Policies}
    \label{tab:ppo_hyperparameters}
    \begin{tabular}{@{}lcc@{}}
        \toprule
        \textbf{Parameter} & \textbf{Arm} & \textbf{Hand} \\
        \midrule
        
        Network Architecture (MLP) & [256, 256, 256] & [256, 256, 256] \\
        Adam Learning Rate & $4 \times 10^{-4}$ & $4 \times 10^{-4}$ \\
        Number of Epochs & 5 & 5 \\
        Number of Mini-batches & 4 & 4 \\
        Discount Factor ($\gamma$) & 0.9751 & 0.9751 \\
        GAE Parameter ($\lambda$) & 0.95 & 0.95 \\
        Clipping Parameter ($\epsilon$) & 0.1214 & 0.1214 \\
        Max Gradient Norm & 1.0 & 1.0 \\
        Desired KL Divergence & 0.01 & 0.01 \\
        
        \bottomrule
    \end{tabular}
\end{table}

\section{Fingertip Jacobian Frame Transformation}
\label{appendix:frame_transformation}

Standard rigid-body dynamics libraries (e.g., Pinocchio~\cite{carpentier2019pinocchio}) typically compute spatial velocities and Jacobians with respect to the global world frame, $\{W\}$. 
However, our hierarchical formulation requires the spatial Jacobian of the fingertip $\{F\}$ relative to the palm $\{P\}$, expressed locally in the palm frame. 
We can derive this relative Jacobian starting from the world-frame quantities.

The spatial velocity of the fingertip in the world frame, ${}^{W}\twist_F$, can be expressed as the sum of the palm's spatial velocity and the fingertip's relative velocity:
\begin{equation}
{}^{W}\twist_F = 
\begin{bmatrix}
\mathbf{I} & \mathbf{0} \\
-[\,{}^{W}\!\mathbf{p}_{F/P}\,]_\times & \mathbf{I}
\end{bmatrix}
{}^{W}\twist_P +
\begin{bmatrix}
{}^{W}\!\mathbf{R}_{P} & \mathbf{0} \\
\mathbf{0} & {}^{W}\!\mathbf{R}_{P}
\end{bmatrix}
{}^{P}\twist_{F/P},
\end{equation}
where ${}^{W}\!\mathbf{p}_{F/P}$ is the position vector from the palm to the fingertip, $[\,\cdot\,]_\times$ denotes the skew-symmetric matrix operator, and ${}^{W}\!\mathbf{R}_{P}$ is the rotation matrix of the palm frame.

Rearranging this expression to solve for the relative spatial velocity ${}^{P}\twist_{F/P}$, and substituting the kinematic mapping $\twist = \jacobian(\genPos)\genVel$, we obtain:
\begin{align} 
{}^{P}\twist_{F/P}
&=
\begin{bmatrix}
{}^{W}\!\mathbf{R}_{P}^\top & \mathbf{0} \\
\mathbf{0} & {}^{W}\!\mathbf{R}_{P}^\top
\end{bmatrix} 
\left( {}^{W}\twist_F - \begin{bmatrix}
\mathbf{I} & \mathbf{0} \\
-[\,{}^{W}\!\mathbf{p}_{F/P}\,]_\times & \mathbf{I}
\end{bmatrix}
{}^{W}\twist_P \right) \nonumber \\
&= 
\begin{bmatrix}
{}^{W}\!\mathbf{R}_{P}^\top & \mathbf{0} \\
\mathbf{0} & {}^{W}\!\mathbf{R}_{P}^\top
\end{bmatrix} 
\left( {}^{W}\!\jacobian_F - \begin{bmatrix}
\mathbf{I} & \mathbf{0} \\
-[\,{}^{W}\!\mathbf{p}_{F/P}\,]_\times & \mathbf{I}
\end{bmatrix}
{}^{W}\!\jacobian_P \right) \genVel \nonumber \\
&=
{}^{P}\!\jacobian_{F/P}(\genPos)\,\genVel.
\end{align}

This derivation reveals a critical kinematic property: the subtraction inside the parentheses perfectly cancels out the contribution of the arm's global motion to the fingertip's relative velocity. 
Consequently, the first seven columns of the resulting relative Jacobian ${}^{P}\!\jacobian_{F/P}(\genPos)$, which correspond to the arm joints, are identically zero. The remaining active columns cleanly isolate the localized contribution of the finger joints.

Finally, we decompose this $6 \times n$ relative spatial Jacobian into its linear and angular blocks:
\begin{equation}
{}^{P}\!\jacobian_{F/P}(\genPos) = 
\begin{bmatrix}
{}^{P}\!\jacobian_{\omega, F/P}(\genPos) \\
{}^{P}\!\jacobian_{v, F/P}(\genPos)
\end{bmatrix}
\end{equation}
In this work, we specifically extract and utilize only the $3 \times n$ translational block, ${}^{P}\!\jacobian_{v, F/P}(\genPos)$, for the fingertip coordination terms in our QP objective (see \autoref{eq:constrained_qp}). 
This design choice reduces the constraint dimensionality, as the desired fingertip orientations are implicitly managed by the hand's mechanical morphology and the high-level planner's reaching strategy.

\section{Computation of Mean Joint Velocity Limit Contours}
\label{appendix:joint_vel_contours}

To compute the mean Cartesian velocity contours constrained by joint velocity limits (shown as black dashed lines in~\autoref{fig:qp_error_spatial}), we evaluate the maximum allowable operational space velocity in a selected 2D plane across a sampled set of configurations.

Let $\mathbf{J}_P \in \mathbb{R}^{6 \times n}$ be the full Jacobian of the robot's palm, where $n$ is the number of joints. For a specific 2D target plane (e.g., $v_x$-$v_y$), we extract the corresponding two rows from the linear portion of the Jacobian to form the planar sub-Jacobian $\mathbf{J}_{xy} \in \mathbb{R}^{2 \times n}$.

For a given angle $\theta \in [0, 2\pi)$, let $\mathbf{u}(\theta) = [\cos\theta, \sin\theta]^\top$ be a 2D Cartesian unit direction vector. For the $k$-th sampled configuration, the minimum-norm joint velocity direction $\delta \genVel^{(k)}(\theta) \in \mathbb{R}^n$ required to move in the direction of $\mathbf{u}(\theta)$ is computed using the Moore-Penrose pseudoinverse $\mathbf{J}_{xy}^{(k)\dagger}$:
\begin{equation}
    \delta \genVel^{(k)}(\theta) = \mathbf{J}_{xy}^{(k)\dagger} \mathbf{u}(\theta)
\end{equation}

Let $\genVel_{\text{max}} \in \mathbb{R}^n$ denote the absolute joint velocity limits. The maximum scalar Cartesian speed $\alpha^{(k)}(\theta)$ along $\mathbf{u}(\theta)$ is constrained by the bottleneck joint. Thus, for the $k$-th sample, the maximum achievable speed is:
\begin{equation}
    \alpha_{\text{max}}^{(k)}(\theta) = \min_{i \in \{1, \dots, n\}} \left( \frac{\genVel_{\text{max}, i}}{\left| \delta \genVel_i^{(k)}(\theta) \right|} \right)
\end{equation}

Given a dataset of $S$ sampled configurations, the mean achievable Cartesian speed in the direction $\theta$ is derived by averaging across all samples:
\begin{equation}
    \bar{\alpha}_{\text{max}}(\theta) = \frac{1}{S} \sum_{k=1}^{S} \alpha_{\text{max}}^{(k)}(\theta)
\end{equation}

Finally, the 2D Cartesian coordinates for the mean joint velocity limit contour, $\mathbf{v}_{\text{limit}}(\theta)$, are obtained by scaling the unit direction vectors by the mean achievable speed:
\begin{equation}
    \mathbf{v}_{\text{limit}}(\theta) = \bar{\alpha}_{\text{max}}(\theta) \mathbf{u}(\theta) = 
    \begin{bmatrix} 
        \bar{\alpha}_{\text{max}}(\theta) \cos\theta \\ 
        \bar{\alpha}_{\text{max}}(\theta) \sin\theta 
    \end{bmatrix}
\end{equation}
Evaluating $\mathbf{v}_{\text{limit}}(\theta)$ over $\theta \in [0, 2\pi)$ yields the closed boundary contours representing the mean physical velocity kinematic limits of the manipulator in the specified Cartesian plane.

}